\documentclass[journal]{IEEEtran}
\usepackage{hyperref}
\hypersetup{
    bookmarks=true,         % show bookmarks bar?
    unicode=false,          % non-Latin characters in Acrobat’s bookmarks
    pdftoolbar=true,        % show Acrobat’s toolbar?
    pdfmenubar=true,        % show Acrobat’s menu?
    pdffitwindow=false,     % window fit to page when opened
    pdftitle={Towards Principled Design of Deep Convolutional Networks: Introducing SimpNet},    % title
    pdfauthor={Seyyed Hossein Hasanpour Matikolaee},     % author
    pdfsubject={CNN Design Principles},   % subject of the document
    pdfcreator={Adobe PDF Creator},   % creator of the document
    pdfproducer={},  % producer of the document
    pdfkeywords={CNN,} {Deep Learning,} {Neural Networks}, % list of keywords
    pdfnewwindow=true,      % links in new window
    colorlinks=false,       % false: boxed links; true: colored links
    linkcolor=red,          % color of internal links
    citecolor=green,        % color of links to bibliography
    filecolor=magenta,      % color of file links
    urlcolor=cyan           % color of external links
}
\usepackage{pdfpages}

\usepackage{subcaption}     % for subfigures
\usepackage[utf8]{inputenc} % allow utf-8 input
\usepackage[T1]{fontenc}    % use 8-bit T1 fonts
\usepackage{hyperref}       % hyperlinks
\usepackage{url}            % simple URL typesetting
\usepackage{booktabs}       % professional-quality tables
\usepackage{amsfonts}       % blackboard math symbols
\usepackage{nicefrac}       % compact symbols for 1/2, etc.
\usepackage{microtype}      % microtypography
\usepackage{epsfig}
\usepackage{graphicx}
\hypersetup{colorlinks=true, linkcolor=red}

\usepackage{algorithm}
\usepackage{algorithmic}
\usepackage{tikz}
\usepackage{pgfplots}
\usepackage{amsmath}
\usepackage{sci}
\usepackage{xfrac}
\usepackage{pgfplots}
\usepackage{pgfplotstable}
\usepackage{filecontents,pgfplots}
\usepackage{tikz}
\usetikzlibrary{arrows,calc,shapes,snakes,positioning,matrix,arrows,decorations.pathmorphing,decorations.text}
\usepackage[sort&compress,square,comma,numbers]{natbib}
\makeatletter
\renewcommand\bibsection%
{
  \section*{\refname
    \@mkboth{\MakeUppercase{\refname}}{\MakeUppercase{\refname}}}
}
\makeatother
\ifCLASSINFOpdf
\else
\fi
\begin{document}

%\title{SimpNet: Simple architectures but better performance}
\title{Towards Principled Design of Deep \\Convolutional Networks: Introducing SimpNet}

\author{Seyyed~Hossein Hasanpour$^*$,
       Mohammad~Rouhani, Mohsen~Fayyaz,
    Mohammad~Sabokrou and Ehsan~Adeli%,~\IEEEmembership{Member,~IEEE,}% <-this % stops a space
\thanks{$^*$Corresponding Author}
\thanks{S.H. Hasanpour is with Arvenware Ltd, Babol, Iran (email: \url{Hossein.Hasanpour@arven.ir})}% <-this % stops a space
\thanks{M. Rouhani is with Technicolor R\&I, Rennes, France (email: \url{Mohammad.Rouhani@technicolor.com})}
\thanks{M. Fayyaz is with University of Bonn, Bonn, Germany (email: \url{fayyaz@iai.uni-bonn.de})}% <-this % stops a space
\thanks{M. Sabokrou is with Institute for Research in Fundamental Sciences (IPM), Tehran, Iran (email: \url{sabokro@ipm.ir})}
\thanks{E. Adeli is with Stanford University, Stanford, CA 94305, USA (email: \url{eadeli@cs.stanford.edu})}
}

\maketitle

\begin{abstract}
Major winning Convolutional Neural Networks (CNNs), such as VGGNet, ResNet, DenseNet, \etc, include tens to hundreds of millions of parameters, which impose considerable computation and memory overheads. This limits their practical usage in training and optimizing for real-world applications. On the contrary, light-weight architectures, such as SqueezeNet, are being proposed to address this issue. However, they mainly suffer from low accuracy, as they have compromised between the processing power and efficiency. These inefficiencies mostly stem from following an ad-hoc designing procedure. In this work, we discuss and propose several crucial design principles for an efficient architecture design and elaborate intuitions concerning different aspects of the design procedure. Furthermore, we introduce a new layer called {\it SAF-pooling} to improve the generalization power of the network while keeping it simple by choosing best features. Based on such principles, we propose a simple architecture called {\it SimpNet}. We empirically show that SimpNet provides a good trade-off between the computation/memory efficiency and the accuracy solely based on these primitive but crucial principles. SimpNet outperforms the deeper and more complex architectures such as VGGNet, ResNet, WideResidualNet \etc, on several well-known benchmarks, while having 2 to 25 times fewer number of parameters and operations. We obtain state-of-the-art results (in terms of a balance between the accuracy and the number of involved parameters) on standard datasets, such as CIFAR10, CIFAR100, MNIST and SVHN. The implementations are available at \href{url}{https://github.com/Coderx7/SimpNet}.
\end{abstract}

\begin{IEEEkeywords}
Deep Learning, Convolutional Neural Networks (CNNS), Simple Network, Classification, Efficiency. 
\end{IEEEkeywords}

\section{Introduction}
\IEEEPARstart{S}{ince} the resurgence of neural networks, deep learning methods have been gaining huge success in diverse fields of applications, including semantic segmentation, classification, object detection, image annotation and natural language processing \cite{Guo_deep_learning_review_2015}. Convolutional Neural Network (CNN), as a powerful tool for representation learning, is able to discover complex structures in the given data and represent them in a hierarchical manner \cite{He_ResNet_2015, Szegedy_googlenet_2015, Simonyan_VGG_2014}. Therefore, there are few parameters to be manually engineered, among which is the network architecture. What all of the recent architectures have in common is the increasing depth and complexity of the network that provides better accuracy for the given task. The winner of the ImageNet Large Scale Visual Recognition Competition 2015 (ILSVRC) \cite{Russakovsky_ImageNet_2015} achieved its success using a very deep architecture of 152 layers \cite{He_ResNet_2015}. The runner up also deployed a deep architecture  \cite{Szegedy_googlenet_2015}. This trend has been followed ever since \cite{Huang_DenselyCNN_2016,Szegedy_inceptiov4_2016,Szegedy_inceptionv3_2016,Zagoruyko_WRN_2016}.

As networks get deeper, aiming to improve their discrimination power, the computations and memory usage cost and overhead become critically expensive, which makes it very hard to apply or expand them for a variety of real-world applications. Despite the existence of various techniques for improving the learning algorithms, such as different initialization algorithms \cite{Glorot_Xavier_Understanding_difficulty_training_dnn_2010, He_PReLU_2015,  Mishkin_AllYouNeedIsGoodInit_2015, Saxe_ExactSolution_2013}, normalization and regularization \cite{Graham_FractionalMaxpooling_2014, Goodfellow_MaxoutNetwork_2013, Ioffe_BatchNorm_incepv2_2015, Wager_dropout_training_2013, Wan_Regularization_Using_DropConnect_2013}, nonlinearities \cite{Clevert_Fast_n_accurat_ELU_2015, He_PReLU_2015, Maas_RectifierNonlinearities_2013, Nair_ReLU_RBM_2010} and data-augmentation tricks \cite{AlexKrizhevsky_imgnet_2012, Graham_FractionalMaxpooling_2014, Simonyan_VGG_2014, Wu_deepImage_2015, Xu_EmpiricalEvalRectified_2015}, they are most beneficial when utilized in an already well performing architecture. In addition, some of these techniques may even impose more computational and memory usage overheads \cite{Goodfellow_MaxoutNetwork_2013, Ioffe_BatchNorm_incepv2_2015,Huang_DenselyCNN_2016}. Therefore, it is highly desirable to design efficient (less complex) architectures with smaller number of layers and parameters, which are as good as their deeper and more complex counterparts. Such architectures can then be further tweaked using novel tricks from the literature.%, and will be easy to improve or fine-tune. 

The key to devising a well performing architecture is to have basic principles that each aim at achieving a specific goal, which together enhance different aspects of a network. Following a principled approach, rather than an ad-hoc one, has the advantage of allowing granular tuning and automated architecture design approaches. In other words, established proven designing principles can be used for achieving a custom architecture that suits a specific goal by the architect or even be made automatically by maneuvering in different design space aspects introduced by such principles. In addition, such maneuvering in the design space effectively exhibits which aspects are more influential and can better result in rectifying existing issues without diverging altogether or working on aspects that are less influential. One of such influential use cases can be the related research on creating an architecture using evolutionary optimization algorithms. With better recognized intuitions, the domain knowledge can be transferred into the algorithm. For instance, Google AutoML aims at automating the design of machine learning models \cite{zoph_automl_learning_2017}.

In this paper, we introduce a set of designing principles to create a road-map for designing efficient networks tailored to the desired goals. We elaborate detailed explanation about different aspects in design, such as different pooling operations, kernel sizes, depth, \etc and provide a good insight into the underlying components in an architecture. Following our defined principles, we propose a new operational layer called ``SAF-pooling'', which enhances the network discrimination power. In this layer, we enforce the network to learn more robust features by first pooling the highest activations, which denote strong features related to the specific class, and then randomly turning off some of them. During this process, we simulate the cases where not all features are present in the input due to occlusion, viewpoint changes, illumination variation and the likes, and thus the network is forced to adapt itself to these new situations by developing new feature detectors. 
%Therefore, by doing such series of operations, in the lower layers of convolutional layers, we increase the generality of features by dropping the important features of an specific sample. However, in the higher layers, each feature map learns more abstracted and class-specific information .Therefore, turning off high activations helps other neurons to learn the class-specific characteristics. \cite{Park_AnalysisOnDropout_2016}.

Considering all these introduced principles, we propose a new simple and efficient architecture denoted as \textit{SimpNet}. To show the effectiveness of such principles, a series of experiments are conducted on 4 major benchmark datasets (CIFAR10/100, SVHN, and MNIST), results of which show that the our architecture outperforms all deeper and heavier architectures, while using 2 to 25 times fewer parameters. Apart from these, each principle is also tested extensively to show their necessity. Simple yet efficient architectures, such as SimpNet, signify the importance of considering such principles. They result in efficient and ideal architectures for many scenarios, especially for deploying in mobile devices (such as drones, cell phones), embedded systems (for Internet of Things, IoT, applications), and in general for Artificial Intelligence (AI) and Deep Learning (DL) applications at the edge (\ie, bringing AI and DL from cloud to the end point devices). They can be further compressed using Compression methods such as \cite{Han_deep_compression_2015, Rastegari_XnorNet_2016} and thus their memory and processing power consumption can be further decreased drastically. We intentionally employ simple and basic methods, in order to avoid any unnecessary cluttered design spaces. Doing this allows us to specifically focus on and identify the main underlying principles and aspects in the design space that greatly influence the performance. Failing to do so brings unnecessary complexity and many side challenges specific to each new method that would ultimately result in either a prolonged process or failing to identify truly important principles. This can ultimately lead to overlooking or mistaking main contributing factors with not-so-relevant criterion, which would be a deviation from the initial goal. Therefore, complementary methods from the literature can, then, be separately investigated. In this way, when the resulting model performs reasonably well, relaxing the constraints on the model (inspired by the above recent complementary methods) can further boost the performance with little to no effort. This performance boost has direct correlation with how well an architecture is designed. A fundamentally clumsily designed architecture would not be able to harness the advantages, because of its inherent \textit{flawed} design.

The rest of the paper is organized as follows: Section \ref{sec:related} presents the most relevant works. In Section \ref{sec:intui}, we present our set of designing principles and in Section \ref{sec:the_arch} we present SimpNet, derived from such principles. In Section \ref{sec:experiments_results} the experimental results are presented on 4 major datasets (CIFAR10, CIFAR100, SVHN, and MNIST), followed by more details on the architecture and different changes pertaining to each dataset. Finally, conclusions and directions for future works are summarized in Section \ref{sec:conclusion}.

\section{Related Works} \label{sec:related}
In this section, we review the latest trends and related works, categorized into two subsections. However, to the best of our knowledge there is no work investigating the network design principles in a general framework. There are previous works such as \cite{He_kaiming_CNN_at_constrained_cost_2015,Szegedy_inceptionv3_2016,Iandola_squeezenet_2016} that proposed some principles or strategies that suited their specific use-cases, although some of them can be applied to any architecture, they were mostly aimed at a specific architecture and therefore did not contain tests or experiments to show how effective they are in other scenarios. Works such as \cite{smith_designpatterns_2016} only listed a few previously used techniques without specifically talking about their effectiveness or validity/usability in a broader sense. It was mere a report on what techniques are being used without any regards on how effective they are in different situations. Strategy-wise, the most similar work to ours is \cite{He_kaiming_CNN_at_constrained_cost_2015}. In the following, we briefly talk about the general trends that have been used in recent years.

\subsection{Complex Networks}
Designing more effective networks were desirable and attempted from the advent of neural networks \cite{Fukushima_Neocognitron_Beginning_1979, Fukushima_Neocognitron_Self_Orgenizing_NN_1980, Ivankhnenko_Polynomial_theory_1971}. This desire manifested itself in the form of creating deeper and more complex architectures \cite{Ciresan_Deep_big_simple_nn_2010, Ciresan_A_committee_of_nn_2011, Ciresan_Multi_colmn_dnn_traffic_sign_2012,AlexKrizhevsky_imgnet_2012, He_ResNet_2015, Huang_DenselyCNN_2016,  Szegedy_googlenet_2015, Simonyan_VGG_2014, Srivastava_HighwayNets_2015, Zagoruyko_WRN_2016}. This was first attempted and popularized by Ciresan \etal~\cite{Ciresan_Deep_big_simple_nn_2010} training a 9 layer multi-layer precenteron (MLP) on a GPU, which was then practiced by other researchers \cite{Ciresan_Deep_big_simple_nn_2010, Ciresan_A_committee_of_nn_2011, Ciresan_Multi_colmn_dnn_traffic_sign_2012, Ciregan_Multi_column_dnn_img_cls_2012, He_ResNet_2015, Huang_DenselyCNN_2016, Nair_ReLU_RBM_2010, Szegedy_googlenet_2015, Simonyan_VGG_2014, Srivastava_HighwayNets_2015, Zagoruyko_WRN_2016}.
Among different works, some played an important role in defining a \textit{defacto} standard in creating and designing architectures. In 2012 Krizhevsky \etal~\cite{AlexKrizhevsky_imgnet_2012} created a deeper version of LeNet5 \cite{Lecun_GradientBased_CNN_1998} with 8 layers called AlexNet, unlike LeNet5, It had a new normalization layer called local contrast normalization, and used rectified linear unit (ReLU) \cite{Nair_ReLU_RBM_2010} nonlinearity instead of the hyperbolic tangent (\ie, $Tanh$), and also a new regularization layer called Dropout \cite{Hinton_preventingCoAdapt_2012}. This architecture achieved state-of-the-art on ILSVRC 2012 \cite{Russakovsky_ImageNet_2015}. %The same year, Le \etal~\cite{Le_Building_HighlevelFeats_2013} trained a gigantic network with 1 billion parameters, which was later proceeded by Coats \etal~\cite{Coates_deepLearning_COTS_HPC_2013} whom trained an 11-billion-parameter network. Both of them were ousted by a much smaller network, AlexNet \cite{AlexKrizhevsky_imgnet_2012}.%
In 2013 Lin \etal~\cite{Lin_NIN_2013} introduced a new concept into the literature and deviated from the previously established trend in the designing networks. They proposed a concept named network-in-network (NIN). they built micro neural networks into convolutional neural networks using $1\times 1$ filters and also used global pooling instead of fully connected layers at the end, acting as a structural regularizer that explicitly enforces feature maps to be confidence maps of concepts and achieved state-of-the-art results on CIFAR10 dataset. In 2014, VGGNet by Simonyan \etal~\cite{Simonyan_VGG_2014} introduced several architectures, with increasing depth from 11 to 19 layers, which denoted deeper ones perform better. They also used $3\times 3$ convolution (conv) filters, and showed that stacking smaller filters results in better nonlinearity and yields better accuracy. In the same year, a NIN inspired architecture named GoogleNet \cite{Szegedy_googlenet_2015} was released, 56 convolutional layers making up a 22 modular layered network. The building block was a block made up of conv layers with $1\times 1$, $3 \times 3$, and $5\times 5$ filters, named an inception module. This allowed to decrease the number of parameters drastically compared to former architectures and yet achieve state-of-the-art in ILSVRC.  In a newer version, each $5 \times 5$ kernel was replaced with two consecutive $3 \times 3$. A technique called batch-normalization \cite{Ioffe_BatchNorm_incepv2_2015} was also incorporated into the network for reducing internal covariate shift, which proved to be an essential part in training deep architectures and achieving state-of-the-art results in the ImageNet challenge.

In 2015, He \etal~\cite{He_PReLU_2015} achieved state-of-the-art results in ILSVRC using VGGNet19 \cite{Simonyan_VGG_2014} with ReLU replaced with a new variant, called Parametric ReLU (PReLU), to improve model fitting. An accompanying new initialization method was also introduced to enhance the new nonlinearity performance. Inter-layer connectivity was a new concept aimed at enhancing the gradient and information flow throughout the network. The concept was introduced by \cite{He_ResNet_2015} and \cite{Srivastava_HighwayNets_2015} independently. In \cite{He_ResNet_2015}, a residual block was introduced, in which layers are let to fit a residual mapping. Accompanied by previous achievements in \cite{He_PReLU_2015}, they could train very deep architectures ranging from 152 to 1000 layers successfully and achieve state-of-the-art on ILSVRC. Similarly, \cite{Srivastava_HighwayNets_2015} introduced a solution inspired by Long Short Term Memory (LSTM) recurrent networks (\ie, adaptive gating units to regulate the information flow throughout the network). They trained 100 to 1000 layer networks successfully.
%Later that year , they proposed a deep architecture of 152 layers, called Residual Network (ResNet) \cite{He_ResNet_2015}, which uses PReLU and a new concept called Residual connections. In ResNet, they used residual blocks, in which layers are let to fit a residual mapping. Training deeper networks became easier and allowed for gaining better accuracy by going deeper without becoming more complex. .
% Huang \etal~\cite{Huang_DeepNN_StochDepth_2016} further enhanced ResNet with stochastic depth, in which a shorter network is trained and then a deeper architecture is used at testing time. They could train even deeper architectures and achieve state-of-the-art on CIFAR dataset. 

%Prior to the ResNet, Srivastava \etal~\cite{Srivastava_HighwayNets_2015} released their Long Short Term Memory (LSTM) recurrent network inspired by highway networks. They used the initialization method proposed by He \etal~\cite{He_PReLU_2015}, and created a special architecture that uses adaptive gating units to regulate the flow of information through the network. They created a 100-layer network, and a 1K-layer network and reported the easy training of such network compared to the plain ones. Their contribution was to show that deeper architectures can indeed be trained with Simple stochastic gradient descent.

In 2016, Szegedy \etal~\cite{Szegedy_inceptiov4_2016} investigated the effectiveness of combining residual connections with their Inception-v3 architecture. They gave empirical evidence that training with residual connections accelerates the training of Inception networks significantly, and reported that residual Inception networks outperform similarly expensive Inception networks by a thin margin. With these variations, the single-frame recognition performance on the ILSVRC 2012 classification task \cite{Russakovsky_ImageNet_2015} improves significantly. % With an ensemble of three residual and one Inception-v4, they achieved 3.08 percent top-5 error on the test set of the ImageNet classification challenge. 
Zagoria \etal~\cite{Zagoruyko_WRN_2016} ran a detailed experiment on residual nets \cite{He_ResNet_2015}  called Wide Residual Net (WRN), where instead of a thin deep network, they increased the width of the network in favor of its depth (\ie, decreased the depth). They showed that the new architecture does not suffer from the diminishing feature reuse problem \cite{Srivastava_HighwayNets_2015} and slow training time. %They report that a 16-layer wide residual network outperforms any previous residual network architectures. They experimented with varying depth of their architecture from 10 to 40 layers, and achieved state-of-the-art result on CIFAR10, 100 and SVHN. 
Huang \etal \cite{Huang_DenselyCNN_2016} introduced a new form of inter-layer connectivity called DenseBlock, in which each layer is directly connected to every other layer in a feed-forward fashion. This connectivity pattern alleviates the vanishing gradient problem and strengthens feature propagation. Despite the increase in connections, it encourages feature reuse and leads to a substantial reduction of parameters. The models tend to generalize surprisingly well, and obtain state-of-the-art in several benchmarks, such as CIFAR10/100 and SVHN.

\subsection{Light Weight Architectures}
Besides more complex networks, some researchers investigated the opposite direction. Springenberg \etal~\cite{Springenberg_StrivingForSimplicity_2014} investigated the effectiveness of simple architectures. They intended to come up with a simplified architecture, not necessarily shallower, that would perform better than, more complex networks. They proposed to use strided convolutions instead of pooling and mentioned that downsampling is enough for achieving good performance, therefore no pooling is necessary. They tested  different versions of their architecture, and using a 17 layer version, they achieved a result very close to the state-of-the-art on CIFAR10 with intense data-augmentation. In 2016, Iandola \etal~\cite{Iandola_squeezenet_2016} proposed a novel architecture called, SqueezeNet, a lightweight CNN architecture that achieves AlexNet-level \cite{AlexKrizhevsky_imgnet_2012} accuracy on ImageNet, with 50 times fewer parameters. They used a previously attempted technique called bottleneck and suggested that the spatial correlation does not matter much and thus $3\times 3$ conv filters can be replaced with $1\times 1$ ones. In another work, \cite{hasanpour_letskeepit_2016} proposed a simple 13-layer architecture, avoiding  excessive depth and large number of parameters, and only utilizing a uniform architecture using $3\times 3$ convolutional and $2 \times 2$ pooling layers. Having 2 to 25 times less parameters, it could outperform much deeper and heavier architectures such as ResNet on CIFAR10, and achieve state-of-the-art result on CIFAR10 without data-augmentation. It also obtained very competitive results on other datasets.
%In this paper, the architecture design is outlined in a more principled way, and a more comprehensive analysis on our newly and more efficiently designed network is provided, both theoretically and experimentally. This analysis is backed with extensive and all-inclusive study of the literature, which pose as evidences for each single principle in our design intuitions.

In 2017, Howard \etal~\cite{howard_mobilenets_2017}, proposed a novel class of architectures with 28 layers called MobileNets, which are based on a streamlined architecture that uses depth-wise separable convolutions to build light weight deep neural networks. 
%They introduced two simple hyperparameters for building mobilenet, which allows for creating the right sized model suited for custom application based on the existing constraints for that application.
They tested their architecture on ImageNet and achieved VGG16 level accuracy while being 32 times smaller and 27 times less computational intensive. Several month later, Zhang \etal~\cite{zhang_shufflenet_2017} proposed their architecture called ShuffleNet, designed specially for mobile devices with very limited computing power (\eg, 10-150 MFLOPs). Using the idea of depth-wise separable convolution, the new architecture proposed two operations, point-wise group convolution and channel shuffle, to greatly reduce computation cost while maintaining accuracy. 
%They could outperform MobileNet under the computation budge of 40 MFLOPs. 

% \subsection{Binarization}
% Hubara \etal~\cite{Hubara_Binarizednn_2016} introduced a method to train Binarized Neural Networks (BNNs), in which during training the binary weights and activations are used for computing the parameter gradients. During the forward pass, BNNs drastically reduce memory size and accesses, and replace most arithmetic operations with bit-wise operations. Rastegari \etal~\cite{Rastegari_XnorNet_2016} also proposed two efficient approximations to standard CNNs: Binary-Weight-Networks and XNOR-Networks. In Binary-Weight-Networks, the filters are approximated with binary values resulting in 32x memory
% saving. In XNOR-Networks, both the filters and the input to conv layers are binary. XNOR-Networks approximate convolutions using primarily binary operations. This results in 58x faster conv operations (in terms of number of the high precision operations) and 32x memory savings.

In this work, we introduce several fundamental designing principles, which can be used in devising efficient architectures. Using these principles, we design a simple 13-layer convolutional network that performs exceptionally well on several highly competitive benchmark datasets and achieves state-of-the-art results in a performance per parameter scheme. The network outperforms nearly all deeper and heavier architectures with several times fewer parameters. 
% in spite of being very simple and not using new tricks in the literature, Our model with several times less number of parameter only performs a tiny bit inferior to the 26M parameter DenseNet (less than 0.37\%). 
The network has much fewer parameters (2 to 25 times less) and computation overhead compared to all previous deeper and more complex architectures. 
%, and performs superior to them despite the huge difference in the number of parameters and depth.
Compared to architectures like SqueezeNet or FitNet (which have less number of parameters than ours while being deeper), our network performs far superior in terms of accuracy. This shows the effectiveness and necessity of the proposed principles in designing new architectures. Such simple architectures can further be enhanced with novel improvements and techniques in the literature. 
%Our model can also be compressed using deep compression techniques to be further enhanced, resulting in a very good candidate for many scenarios. Such achievements prove the usability of the proposed network design intuitions which can also help other researchers to design better network architectures with less effort. 

\section{Design Intuitions} \label{sec:intui}
Designing an optimal architecture requires a careful consideration and compromise between the computational complexity and the overall performance of the framework. There are several different factors contributing to each of these two. Previous works often neglected one or more of such factors based on their applications. Here, we study these factors, and clearly categorize and itemize them as principles to take into consideration when designing a network. Then, based on these principles we propose an architecture and illustrate that we can obtain comparable or even better results, while requiring much less computational resources.

\subsection{Gradual Expansion with Minimum Allocation}\label{sec:gradual_expansion_n_minumum_alloc}
% Start with a small network, expand it into a deep but thinner network gradually. Not the number of depth nor the number of parameters are good indicators of how a network should perform. They are neutral factors that are only beneficial when utilized mindfully, if enough care is not taken, the design would be an inefficient network imposing un-wanted overhead. It is also important to note that, a specific depth does not necessarily work for all problems. 
%A brief discussion about depth vs width, and their importance : 
The very first thought striking to mind, when designing a network, is that it should be a very deep architecture. It is a widely accepted belief in the literature that deeper networks perform better than their shallower counterparts. One reason for such a wide spread belief is the success of deeper models in recent years~\cite{Eigen_Lecun_Understanding_deepArch_recurs_cnn_2013, He_ResNet_2015, Simonyan_VGG_2014, Szegedy_googlenet_2015, Geras_Blending_Lstm_into_CNN_2015, Larochelle_empiricaleval_deepnn_2007, Seide_ConversationalSpecch_2011, Srivastava_TrainingVeryDeepNets_highway_2015, Urban_doDeepCNNNeedBeDeep_2016}. Romero \etal~\cite{Romero_Fitnet_2014} also showed how a deeper and thinner architecture performs better. It makes sense that by adding more layers, we are basically providing more capability of learning various concepts and relations between them. Furthermore, it is shown that early layers learn lower level features while deeper ones learn more abstract and domain specific concepts. Therefore, a deeper hierarchy of such features would yield better results \cite{AlexKrizhevsky_imgnet_2012, Zeiler_VisualizingCNN_2014}. 
%something that can indeed be seen in recent years. Moreover, As stated by \cite{Larochelle_empiricaleval_deepnn_2007} and also verified by ImageNet competition \cite{JiaDeng_FeiFeiLi_ImageNet_2009} results for the past several years~\cite{AlexKrizhevsky_imgnet_2012, He_ResNet_2015, Simonyan_VGG_2014, Szegedy_googlenet_2015},
Furthermore, More complex datasets seem to benefit more from deeper networks, while simpler datasets work better with {\it relatively} shallower ones \cite{Larochelle_empiricaleval_deepnn_2007,AlexKrizhevsky_imgnet_2012, He_ResNet_2015, Simonyan_VGG_2014, Szegedy_googlenet_2015}. However, while deeper architectures do provide better accuracy compared to a shallower counterpart, after certain depth, their performance starts to degrade \cite{He_kaiming_CNN_at_constrained_cost_2015}, and they perform inferior to their shallower counterpart, indicating a shallower architecture may be a better choice. Therefore, there have been attempts to show that wider and shallower networks can perform like a deep one \cite{hasanpour_letskeepit_2016,Zagoruyko_WRN_2016}. The depth of the network has a higher priority among other aspects in achieving better performance. Based on these results, it seems a certain depth is crucial to attain a satisfactory result. However, the ratio between depth, width and number of parameters are unknown, and there are conflicting results against and in support of depth. 
%Actual reasoning starts from here:
Based on the above discussions, to utilize depth, memory and parameters more efficiently, it is better to design the architecture in a gradual fashion, \ie, instead of creating a network with a random yet great depth, and random number of neurons, beginning with a small and thin network then gradually proceeding with deepening and then widening the network is recommended. This helps to prevent excessive allocation of processing units, which imposes unnecessary overhead and causes overfitting. Furthermore, employing a gradual strategy helps in managing how much entropy a network would provide. The more parameters a network withholds, the faster it can converge and the more accuracy it can achieve, However it will also overfit more easily. A model with fewer parameters, which provides better results or performs comparable to its heavier counterpart indicates the network has learned better features for the task. In other words, by imposing more constraints on the entropy of a network, the network is implicitly forced to find and learn much better and more robust features. This specifically manifests itself in the generalization power, since the network decisions are based on more important, more discriminative, and less noisy features. 
%Depth and width discussion continues here : 
By allocating enough capacity to the network, a shallower and wider network can perform much better than when it is {\it randomly} made deeper and thinner, as shown in our experiments and also pointed out by~\cite{He_kaiming_CNN_at_constrained_cost_2015, Szegedy_inceptionv3_2016, Zagoruyko_WRN_2016}. It is also shown that widening an existing residual architecture significantly improves its performance compared to making it deeper \cite{Zagoruyko_WRN_2016}. Thus, instead of going any deeper with thinner layers, a wider and comparatively shallower "but still deep enough" is a better choice. Additionally, it is computationally more effective to widen the layers, rather than having thousands of small kernels, as GPUs are often much more efficient in parallel computations on large tensors \cite{Zagoruyko_WRN_2016}. %Choosing a very deep network to begin with is not recommended, because this usually results in more memory usage and in order to keep both memory and computation overhead in check, its best to follow a gradual approach. 

Failing to give enough care at this stage will give rise to issues concerning very deep architectures (such as weakened gradient flow, degradation issue \cite{He_ResNet_2015} and computation usage overhead), while it is usually unnecessary to maintain such a depth in many applications. It is recommended to expand the network to reach a pyramid-shaped form, which means a progressive reduction of spatial resolution of the learned feature maps with the increase in the number of feature maps to keep the representational expressiveness. A Large degree of invariance to geometric transformations of the input can be achieved with this gradual reduction of spatial resolution compensated by a progressive increase of the richness of the representation (the number of feature maps) \cite{Lecun_GradientBased_CNN_1998}.
It is important to note that, one may find out that with the same number of parameters, a deeper version might not achieve as good accuracy as a shallower counterpart and this looks like a contradiction to what we have previously discussed. The improved performance in the shallower architecture can be attributed to the better allocation of layers processing capacity. Each layer in an architecture needs a specific number of processing units in order to be able to carry out its underlying task properly. Hence, with the exact same number of parameters, these parameters will be scattered among the shallower layers better than a much deeper architecture. It is evident that in the deeper counterpart, with the same processing budget, these fewer units in each layer will bear less processing capacity and, hence, a decreased and degraded performance will result. In such a circumstance, we can say that we have an underfitting or low-capacity issue at the layer level. The decreased processing capacity will not let the network take advantage of the available depth, and hence the network is unable to perform decently. A much deeper architecture also exhibits a higher chance of ill-distribution of processing units to the other extreme.
Furthermore, properly distributing neurons between a shallower and a deeper architecture, using the same processing budget, is harder for the deeper counter part. This issue even gets more pronounced, when the architecture is made further deeper. In other words, as the difference between the depth of the two increases, the job of properly allocating neurons to all layers becomes even harder and ultimately at some point will be impossible with the given budget. When a deeper architecture is deprived of the needed processing capacity, a Processing Level Deprivation (PLD) phenomena occurs, in which the architecture fails to develop simple but necessary functions to represent the data. %This can be specifically attributed to the linear nature of convolution operation (\ie, to account for its inherent linearity shortcoming, more filters need to be developed to enable the separation of a nonlinear latent concept in the input).
This causes the deeper network to perform inferior to the shallower one. That is one of the main issues that arises when a shallower and a deeper architecture are being compared performance-wise. Accordingly, when the processing budget increases, the shallower architecture starts to perform inferior to the deeper counterpart, and this gets more pronounced as the budget is further increased. This is because the shallower architecture has saturated and fails to properly utilize the existing capacity, and thus a phenomena called Processing Level Saturation (PLS) occurs, where more processing power would not yield in increased representational power. Meanwhile, the deeper architecture can now utilize the increased processing power and develop more interesting functions, resulting in PLD vanishing and further improvements in model performance. As the number of parameters increases, this difference is even more vigorously noticed. This is in turn one of the reasons, deeper architectures are usually larger in the number of parameters (compared to shallower ones). Note that if we consider an increased budget (suited for a deeper architecture so that all layers can be capacitated properly to carry on their task), the same number of parameters will over-saturate the shallower network and result only in unnecessary and wasted processing power (and vice versa). This also shows why gradual expansion and minimum allocation are key concepts, as they prevent first from choosing a very deep architecture and second from allocating too much neurons, thus, preventing the PLD and PLS issues effectively.

\subsection{Homogeneous Groups of Layers} 
The design process of typical architectures has conventionally been treated as simply putting together a stack of several types of layers such as Convolution, Pooling, Normalization and the likes. Instead of viewing the design process as a simple process of stacking a series of individual layers, it is better to thoughtfully design the architecture in groups of homogeneous layers. The idea is to have several homogeneous groups of layers, each responsible for a specific task (achieving a single goal). 
This symmetric and homogeneous design, not only allows to easily manage the number of parameters a network will withhold while providing better information pools for each semantic level, but will it also provide the possibility of further granular fine-tuning and inspection, in a group-wise fashion. This technique has been in use implicitly, since the success of \cite{AlexKrizhevsky_imgnet_2012} and is also being used by all recent major architectures more profoundly (accentuated) than before, including \cite{He_ResNet_2015, Iandola_Densenet_efficient_2014, Szegedy_inceptiov4_2016, Szegedy_inceptionv3_2016, Zagoruyko_WRN_2016,Huang_DenselyCNN_2016}. However, the previous use of this technique has been only to showcase a new concept rather than to fully take advantage of such building block. Therefore, other aspects and advantages of such feature has not yet been fully harnessed. In other words, almost all former use-cases have been following an ad-hoc formation \cite{Szegedy_inceptiov4_2016} in utilizing such concept inside the network. Nevertheless, this scheme greatly helps in managing network topology requirements and parameters.

\subsection{Local Correlation Preservation}
It is very important to preserve locality throughout the network, especially by avoiding $1\times1$ kernels in early layers. The corner stone of CNN success lies in the local correlation preservation \cite{Lecun_GradientBased_CNN_1998, lecun_nature_2015}. However, \cite{Iandola_squeezenet_2016} has a contrary idea and reported that using more $1\times1$ (as opposed to more $3\times3$) in their architecture had a better result, and thus spatial resolution in CNN is not as important as it may have looked. Our extensive experiments along with others \cite{Simonyan_VGG_2014,Szegedy_inceptionv3_2016,Lecun_GradientBased_CNN_1998} show the contrary and we argue this is the result of following an ad-hoc procedure in evaluating the hypothesis and more thorough experiments could yield a different outcome. For instance, the distribution of filters throughout the network does not follow a principled strategy, the same thing applies to the ad-hoc creation of the architecture, which would prevent the network from fully utilizing the capacity provided by bigger kernels. One reason can be the excessive use of bottlenecks to compress the representations that yielded such results. Further experiments showed indications of our claim, where a much shallower architecture utilizing only $3\times3$ kernels with fewer parameters could outperform SqueezeNet with twice the number of parameters. %Spatial resolution in CNN is indeed a very important factor. %As it can further be seen by our architecture being vastly superior to them with much fewer number of parameters. (SqueezeNet 1.1 with 726K parameters achieves $88.60(by being optimized by us,achieved 92.20)\%$ accuracy on cifar10, while our 300K achieves $93.25\%$).  
Based on our extensive experiments, it is recommended not to use $1\times 1$ filters or fully connected layers, where locality of information matters the most. This is exclusively crucial for the early layers in the network. $1\times 1$ kernels bear several desirable characteristics, such as increasing networks nonlinearity and feature fusion \cite{Lin_NIN_2013} and also decreasing number of parameters by factorization and similar techniques \cite{Szegedy_inceptionv3_2016}. However, on the other hand, they ignore any local correlation in the input. Since they do not consider any neighborhood in the input and only take channel information into account, they distort valuable local information. Nevertheless, the $1\times 1$ kernels are preferable in the later layers of the network. It is suggested to replace $1\times 1$ filters with $2\times 2$ if one plans on using them in places other than the ending layers of the network. Using $2\times 2$ filters reduces the number of parameters, while retaining the neighborhood information. %In case a feature vector is available as apposed to a feature-map, it is still recommend to use conv1D and still try to preserve correlation.
It should also be noted that excessive shrinkage in bottleneck strategy using $1\times 1$ filters should be avoided or it will harm the representational expressiveness \cite{Szegedy_inceptionv3_2016}. 

While techniques, such as aggregation and factorization heavily utilized in \cite{Szegedy_inceptionv3_2016}, provide a good way for replacing larger kernels by smaller ones and decreasing computation massively, we specifically do not use them for several reasons: (1) To intentionally keep everything as simple as possible, evaluate the basic elements, and identify crucial ones, which can then be incorporated in other schemes including inception-like modules. (2) Trying to replace all kernels into a sequence of smaller ones without knowing their effectiveness would increase the depth of the network unnecessarily (\eg, one $5\times 5$ would need two $3 \times 3$ or four $2 \times 2$ kernels), and would therefore quickly provide difficulties in managing the network efficiency. (3) Knowing the effectiveness of different kernel sizes helps utilizing this technique judiciously, and thus, improve the network performance. (4) They have been used in multi-path designs such as inception \cite{Szegedy_googlenet_2015,Szegedy_inceptionv3_2016,Szegedy_inceptiov4_2016}, where several convolutional layers are concatenated with various filter sizes or numbers, and there is not enough evidence of how they work individually \cite{He_kaiming_CNN_at_constrained_cost_2015}. It seems they work well in cases where complementary kernels are also being incorporated alongside. So, their effectiveness in isolation requires more research and experimentation. Furthermore, the principle of choosing hyperparameters concerning such multi-path designs requires further investigation. As an example, the influence of each branch remains unclear \cite{He_kaiming_CNN_at_constrained_cost_2015}, and therefore we only consider ``single-path'' designs with no parallel convolutional layers (already faced with abundance of choices). (5) While modules such as inception do reduce the number of parameters \cite{Szegedy_inceptionv3_2016}, they impose computation overhead and their complex nature makes it even harder to customize it for arbitrary scenarios, since there are many variables to account for. Therefore, we resorted to single path design. At the very minimum, using these principles improves the results. 

Finally It is noteworthy to avoid shrinking feature-map size at ending layers too much and at the same time, allocate excessive amount of neurons to them, while specifically earlier layers have much fewer numbers. very small feature-map sizes at the end of the network (\eg, $1 \times 1$) leads to a minuscule information gain %due to the absence of sufficient neighborhood information, 
and allocating too much neuron would only result in wasted capacity.

\subsection{Maximum Information Utilization}
It is very important to avoid rapid downsampling or pooling, especially in early layers. Similar suggestion has also been given by \cite{He_kaiming_CNN_at_constrained_cost_2015, Iandola_squeezenet_2016, Szegedy_inceptionv3_2016}. To increase the network's discriminability power, more information needs to be made available. This can be achieved either by a larger dataset (\ie, collecting more data, or using augmentation) or utilizing available information more efficiently in the form of larger feature-maps and better information pools. Techniques, such as Inception \cite{Szegedy_googlenet_2015,Ioffe_BatchNorm_incepv2_2015,Szegedy_inceptiov4_2016,Szegedy_inceptionv3_2016}, inter layer connectivity including residual connections \cite{He_ResNet_2015}, Dense connections \cite{Huang_DenselyCNN_2016}, and pooling fusion \cite{Lee_CNN_Mixed_gated_2016}, are examples of more complex ways of providing better information pools. However, at the very least in its most basic form, an architecture can achieve better information pool without such complex techniques as well. If larger dataset is not available or feasible, the existing training samples must be efficiently utilized. 
%(although failing to do so in first place would waste a lot of information provided by using a larger dataset as well, so in any case this is helpful).
Larger feature-maps, especially in the early layers, provide more valuable information in the network compared to the smaller ones. With the same depth and number of parameters, a network that utilizes larger feature-maps achieves a higher accuracy ~\cite{He_kaiming_CNN_at_constrained_cost_2015, Simonyan_VGG_2014, Szegedy_inceptionv3_2016}. Therefore, instead of increasing the complexity of a network by increasing its depth and its number of parameters, one can leverage better results by simply using larger input dimensions or avoiding rapid early downsampling. This is a good technique to keep the complexity of the network in check and to increase the accuracy. Similar observation has been reported by \cite{Simonyan_VGG_2014,Szegedy_googlenet_2015,Szegedy_inceptionv3_2016} as well. Utilizing larger feature-maps to create more information pools can result in large memory consumption despite having a small number of parameters. A good example of such practice can be observed in DenseNet \cite{Huang_DenselyCNN_2016}, from which a model (DenseNet-BC, $L=100$, $k=12$, with $0.8M$ parameters) takes more than 8 gigabytes of video memory (or VRAM). Although such memory consumption in part can be attributed to the inefficient implementation, still a great deal of that is present even after intensive optimization. Other instances can be seen in \cite{He_ResNet_2015} and \cite{Zagoruyko_WRN_2016} that utilize many layers with large feature-maps. This shows that part of these architectures success is because of the information pool that is being provided for each semantic level using an increased number of layers with feature maps of the same size (without being downsampled). Moreover, other than data-augmentation, which has the most influential role in achieving translation invariance \cite{Kauderer_QuantifyingTranslation_invar_2016}, (max-)pooling plays an important role in reducing the sensitivity of the output to shift and distortions \cite{Lecun_GradientBased_CNN_1998}. It also helps achieving translation invariance to some extend \cite{He_kaiming_CNN_at_constrained_cost_2015, Scherer_EvaluationOfPooling_2010}. Scherner \etal~\cite{Scherer_EvaluationOfPooling_2010} show that max-pooling operation is vastly superior for capturing invariances in image-like data, compared to a subsampling operations. Therefore, it is important to have reasonably enough number of pooling layers in the architecture. Since the use of max-pooling will distort the exact location, excessive appliance of such layer would negatively impact the performance in some applications, such as semantic segmentation and object detection. Proper application of pooling both results in obtaining translation invariance, and imposes less memory and computation overheads. It also needs to be noted that the pooling operation, in its essence is different than simple downsampling. %Although a pooling layer has down-sampling intermingled with pooling operation, it doesn't mean its effective outcome can be achieved using a simple down-sampling operation.
It is usually theorized that a convolutional layer can ``learn'' the pooling operation better. The notion that a pooling operation can be learned in a network is correct and have already been tried many times \cite{Lee_CNN_Mixed_gated_2016,Goodfellow_MaxoutNetwork_2013,gulcehre_earned-norm_pooling_2014,jia_beyondspatialpyramids_2012}, however, examples such as strided convolutions \cite{Springenberg_StrivingForSimplicity_2014} which are convolutions with bigger strides, are far from doing that. %approximating or simulating a pooling operation. 
Techniques such as strided convolution thus do not yield the same effect as a pooling operation, since a simple convolution layer with rectified linear activation cannot by itself implement a p-norm computation \cite{Springenberg_StrivingForSimplicity_2014}. Furthermore It is argued \cite{Springenberg_StrivingForSimplicity_2014} that the improvement caused by pooling is solely because of the downsampling and thus, one can replace a pooling operation with a strided convolution. However, our empirical results and further intuitions, which result in improved performance by pooling (explained later), prove otherwise. This is more visible in our tests, where with the same number of parameters, the architecture with simple max-pooling always outperforms the strided convolution counterpart. This is thoroughly addressed in the experiments section. 
%As it is shown in our experiments, replacing strided convolutions with pooling or increasing the network parameter which uses pooling to match the number of parameters of the strided version, clearly shows the pooling operation outperforms the strided version, proving our claim. Thus it is important to have pooling operation (not only a down-sampling) since it actually is a mean to capture prominent features from former layers and thus can (if properly applied) speed convergence and network generalization.

\begin{figure}[t]
\centering
\begin{tikzpicture}
    \begin{axis}[
        width  = 0.4\textwidth,
        height = 3.2cm,
        major x tick style = transparent,
        ybar=2\pgflinewidth,
        bar width=5pt,
        ymajorgrids = true,
        ylabel = {\footnotesize Relative training time},
        ylabel style={yshift=-0.6cm},
        symbolic x coords={Alexnet,Googlenet,VGG},
        xtick = data,
        scaled y ticks = false,
        enlarge x limits=0.25,
        ymin=0,
        legend cell align=left,
                legend style={
                    draw=none, % ?
                    text depth=0pt,
                    at={(-0.1,1.38)},
                    anchor=north west,
                    legend columns=2,
                    % default spacing:
                    column sep=0.2cm,
                    % The text "Legend:"
                    %/tikz/column 2/.style={column sep=0pt,font=\bfseries},
                    %
                    % the space between legend image and text:
                    /tikz/every odd column/.append style={column sep=0cm},
                },
    ]
        \addplot[style={bblue,fill=bblue,mark=none}, area legend]
            coordinates {(Alexnet, 1) (Googlenet,1) (VGG,1)};

        \addplot[style={rred,fill=rred,mark=none}, area legend]
             coordinates {(Alexnet,2.2) (Googlenet,2.2) (VGG,2.9)};
        \legend{cuDNN v4+K40,cuDNN v5.1+M40}
    \end{axis}
\end{tikzpicture}
\caption{Comparisons of the speedup of different architectures. This plot shows $2.7\times$ faster training when using $3\times3$ kernels using cuDNN v5.x.}
    \label{fig:cud}
\end{figure}
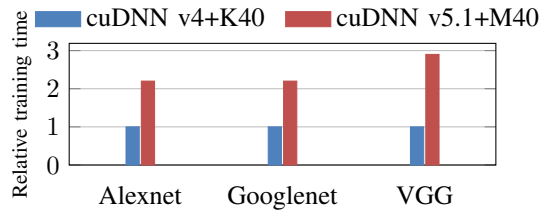
\begin{table}[t]
    \caption{Improved performance by utilizing cuDNN v7.x.}
    \label{tab:cuDNNv7}
    \centering
    \begin{tabular}{cccc}
       & k80+cuDNN 6 & P100+ cuDNN 6 & v100+cuDNN 7 \\ \hline
    2.5x + CNN & 100 & 200 &600 \\
        3x + LSTM  & 1x & 2x & 6x \\ \hline
    \end{tabular}
\end{table}
%\vspace{-3mm}

\subsection{Maximum Performance Utilization}
%Use 3x3, and follow established industrial trends. For an architecture to be easily usable and widely practical, it needs to perform fast and decently. 
Considering implementation details and recent improvements in the underlying libraries, one can simply design better performing and more efficient architectures. For instance, using $3\times 3$ filters, besides its already known benefits ~\cite{Simonyan_VGG_2014}, allows to achieve a substantial boost in performance, when using NVIDIA's Cudnnv5.x library or higher. This is a speed up of about $2.7\times$ compared to the former v4 version. The performance boost can be witnessed in newer versions such as Cudnnv7.x as well. This is illustrated in Figure \ref{fig:cud} and Table \ref{tab:cuDNNv7}. The ability to harness every amount of performance is crucial, when it comes to production and industry. Whereas, using larger kernels such as $5\times 5$ and $7\times7$ tend to be disproportionally more expensive in terms of computation. For example, a $5 \times 5$ convolution with $n$ filters over a grid with $m$ filters is $\sfrac{25}{9} = 2.78$ times more computationally expensive than a $3 \times 3$ convolution with the same number of filters. Of course, a $5\times 5$ filter can capture dependencies between signals activation of units further away in the earlier layers, so a reduction of the geometric size of the filters comes at a larger cost \cite{Szegedy_inceptionv3_2016}. Besides the performance point of view, on one hand larger kernels do not provide the same efficiency per parameter as a $3 \times 3$ kernel does. It may be interpreted that since larger kernels capture a larger area of neighborhood in the input, they may help suppressing noise to some extend and thus capturing better features. However, in practice the overhead they impose in addition to the loss in information they cause makes them not an ideal choice. Furthermore, larger kernels can be factorized into smaller ones \cite{Szegedy_inceptionv3_2016}, and therefore using them makes the efficiency per parameter to decline, causing unnecessary computational burden. Substituting larger kernels with smaller ones is also previously investigated in \cite{Szegedy_inceptiov4_2016, Szegedy_inceptionv3_2016}. Smaller kernels, on the other hand, do not capture local correlations as well as $3\times3$ kernels. Detecting boundaries and orientations are better done using a $3\times3$ kernel in earlier layers. A cascade of $3\times3$ can also replace any larger one and yet achieve similar effective receptive field. In addition, as discussed, they lead to better performance gain. 

\subsection{Balanced Distribution Scheme} \label{sec:balanced_distribution}
Typically, allocating neurons disproportionally throughout the network is not recommended, because the network would face information shortage in some layers and operate ineffectively. For example, if not enough capacity is given to early layers to provide low-level features, or middle layers to provide middle-level features, the final layers would not have access to sufficient information to build on. This applies to all semantic levels in a network, \ie, if there is not enough capacity in the final layers, the network cannot provide higher level abstractions needed for accurate deduction. %If middle layers fail to acquire needed capacity, they won’t be able to provide needed data for upper layer(s) and upper layer(s) will not be able to efficiently utilize their processing capacity.
Therefore, to increase the network capacity (\ie, number of neurons) it is best to distribute the neurons throughout the whole network, rather than just fattening one or several specific layers \cite{Szegedy_inceptionv3_2016}. %Spread the neurons according to the principles noted before. That is, 
The degradation problem that occurs in deep architectures stems from several causes, including the ill-distributed neurons. As the network is deepened, properly distributing the processing capacity throughout the network becomes harder and this often is the cause for some deeper architectures under-performing against their shallower counterparts. This is actually what arises to PLD and PLS issues. we address this in the experiments section in details. 
%As a result, it is better to allocate neurons to all layers proportionally. This means that as we approach the end of the network, more neurons should be allocated to each layer. In other words, earlier layers need to have properly been capacitated.
Using this scheme, all semantic levels in the network will have increased capacity and will contribute accordingly to the performance, whereas the other way around will only increase some specific levels of capacity, which will most probably be wasted.

\subsection{Rapid Prototyping In Isolation}
It is very beneficial to test the architecture with different learning policies before altering it. Most of the times, it is not the architecture that needs to be changed, rather it is the optimization policy that does. A badly chosen optimization policy leads to inefficient convergence, wasting network resources. Simple things, such as learning rates and regularization methods, usually have an adverse effect if not tuned correctly. Therefore, it is first suggested to use an automated optimization policy to run quick tests and when the architecture is finalized, the optimization policy is carefully tuned to maximize network performance. %we have some images that show different acc using different policies (with the same architecture) due to shortage of space I didnt add them here , I dont know if that was a good idea!.
It is essential to note that when testing a new feature or applying a change, everything else (\ie, all other settings) remain unchanged throughout the whole experimental round. For example, when testing $5\times 5$ \vs~$3\times 3$, the overall network entropy must remain the same. It is usually neglected in different experiments and features are not tested in isolation or better said, under a fair and equal condition which ultimately results in a not-accurate or, worse, a wrong deduction. %Throughout experimentation, it is easy to slip and lose track of the changes, some architectures contain more parameters and entropy, some might become deeper, \etc~%In order to effectively assess a specific feature and its effectiveness in the architectural design, it is important to keep track of changes the specific feature causes and take necessary action(s) to provide the required environment to conduct the test.

\subsection{Dropout Utilization}
\begin{figure}[t]
\begin{center}
\begin{tikzpicture}
  \begin{axis}[width=9cm, height=5cm,
    symbolic x coords = {1, 2, 3, 4, 5,6,7,8,9,10,11,12,13,14,15,16,17,18,19,20,21,22,23,24,25,26,27,28,29,30,31,32,33,34,35,36,37,38,39,40,41,42,43,44,45,46,47,48,49,50,51,52,53,54,55,56,57,58,59,60,61,62,63,64,65,66,67},
    legend pos = north east,
  ]
  \addplot+[smooth] coordinates { (2,28.691518)(3,26.19039154
)(4,16.75759888
)(5,26.38938522)(6,26.881774)(7,12.68436146)(8,26.48947334)(9,24.84106064)(10,26.7809906)(11,19.64202118)(12,28.34988403)(13,23.15718079)(14,26.2544651)(15,26.75907516)(16,30.94452286)(17,17.75754929)(18,30.46988678)(19,25.72103119)(20,25.94272614)(21,30.30309486)(22,20.96222496)(23,17.20010185)(24,7.362791061)(25,17.70526695)(26,27.70537949)(27,22.2742424)(28,20.97610474)(29,20.16061401)(30,33.8180275)(31,30.2789917)(32,25.02797127)(33,61.08970261)(34,36.5642128)(35,15.16472244)(36,20.65788078)(37,24.58080292)(38,34.70709991)(39,20.1482048)(40,25.42594147)(41,28.29319382)(42,29.10025215)
(43,25.12285614)(44,27.94253159)(45,23.50538445)(46,42.86722565)(47,21.69746399)
(48,30.90040588)(49,27.30561256)(50,14.73661518)(51,29.5870285)(52,11.42878532)
(53,21.54691696)(54,27.12130356)(55,27.90488625)(56,27.8960495)(57,23.00183296)(58,18.75677299)(59,18.25593567)(60,18.46562195)(61,32.92595673)(62,27.87899399)(63,22.32618332)(64,27.03840828)(65,22.58941269)(66,17.95868874)(67,26.03449249)
};
\addplot+[smooth] coordinates { (2,71.9921875
)(3,1.853786707)(4,0)(5,29.01682281)(6,0.289011031)(7,6.781476974)(8,18.0578537)(9,0)(10,0)(11,56.74181366)(12,44.45471191)(13,0.559480846)(14,0)(15,0)(16,0)(17,5.930178642)(18,0)(19,12.74249077)(20,0)(21,6.605689049)(22,35.86524963)(23,19.44761658)(24,6.78802681)(25,7.886648655)(26,0.626071811)(27,106.4539185)(28,33.13948822)(29,1.744260311)(30,0)(31,4.888971329)(32,0)(33,0)(34,0.095655002)(35,0)(36,20.50197983
)(37,28.1124382)(38,28.50383377)(39,0.175519079)(40,48.96050262)(41,1.573083282)(42,19.33377457)(43,11.09307289)(44,0)(45,34.23406982)(46,0.069998533)(47,29.24234009)
(48,0.071755081)(49,45.18746185)(50,0.403084755)(51,0.332451463)(52,0)(53,0.206146196)(54,19.97655106)(55,36.16522598)(56,0)(57,0.252303034)(58,0)(59,0)(60,1.01092875
)(61,4.570125103)(62,0)(63,20.24185944)(64,0.132169887)(65,0)(66,0.213888973)(67,48.72380447)};
  
  \legend{{\small With dropout},{\small Without dropout}}
  \end{axis}
\end{tikzpicture}
\end{center}
   \caption{Layer 1 neuron activations with and without dropout. More neurons are frequently activated when dropout is used, which means more filters are better developed.}
\label{fig:drpout_vs_nodrp_activations_firstlayer}
\end{figure}
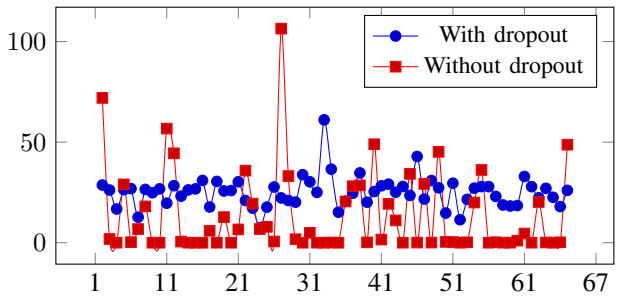

\begin{figure}[h!]
  \centering
  \begin{subfigure}[b]{0.49\linewidth}
    \includegraphics[width=\linewidth]{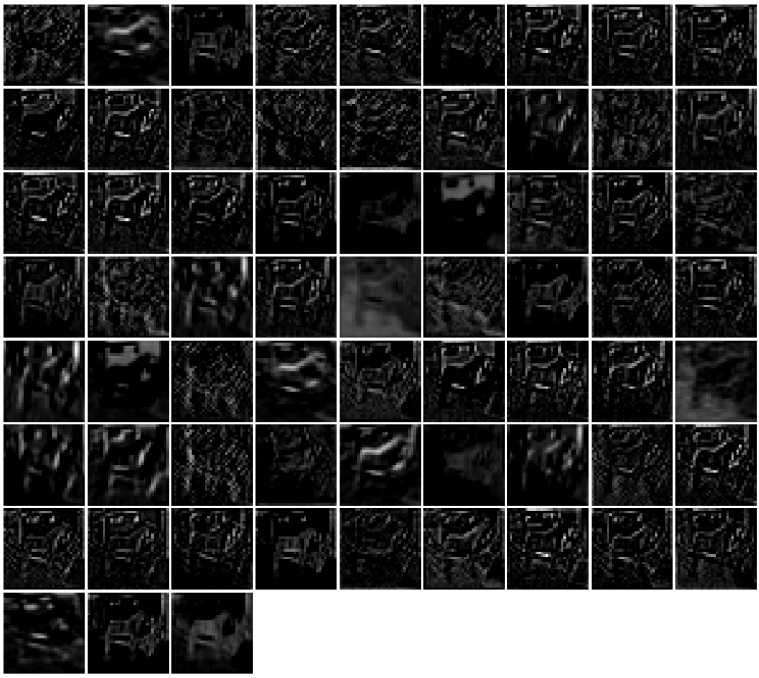}
    \caption{With dropout.}
  \end{subfigure}
  \begin{subfigure}[b]{0.49\linewidth}
    \includegraphics[width=\linewidth]{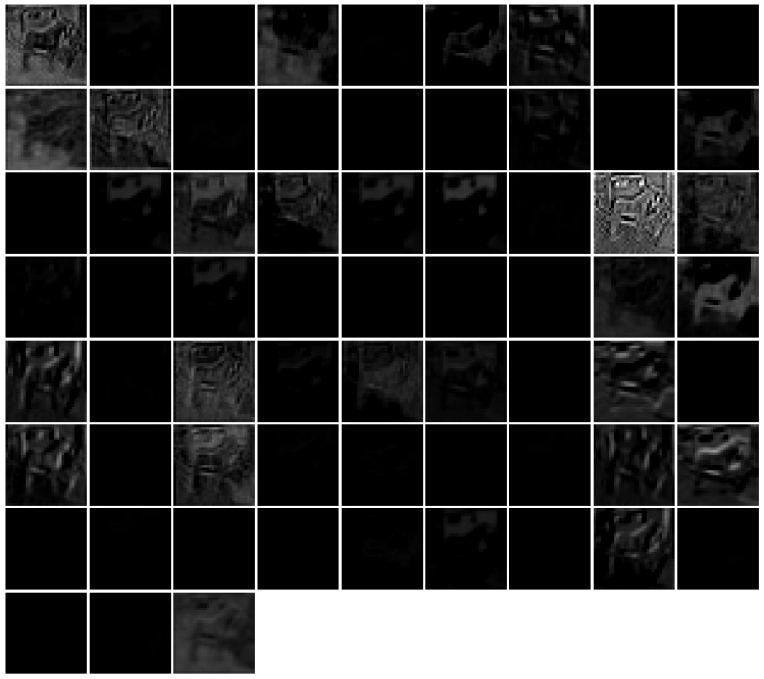}
    \caption{Without dropout.}
  \end{subfigure}
  \caption{When dropout is used, all neurons are forced to take part and thus less dead neurons occur. It also results in the development of more robust and diverse filters.}
  \label{fig:new_conv1_fmap}
\end{figure}

\begin{figure}[h!]
  \centering
  \begin{subfigure}[b]{0.48\linewidth}
    \includegraphics[width=\linewidth]{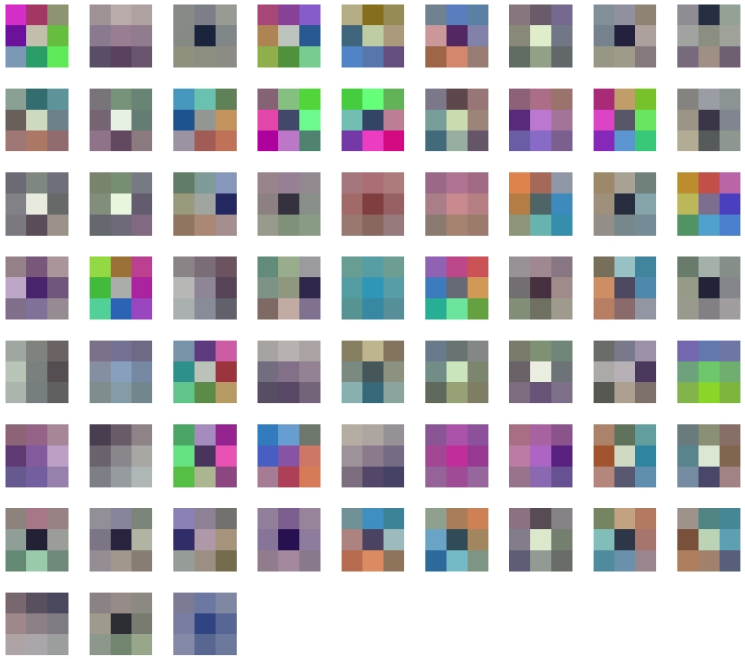}
    \caption{Filters learned in the first layer, when dropout is used.}
  \end{subfigure} ~
  \begin{subfigure}[b]{0.48\linewidth}
    \includegraphics[width=\linewidth]{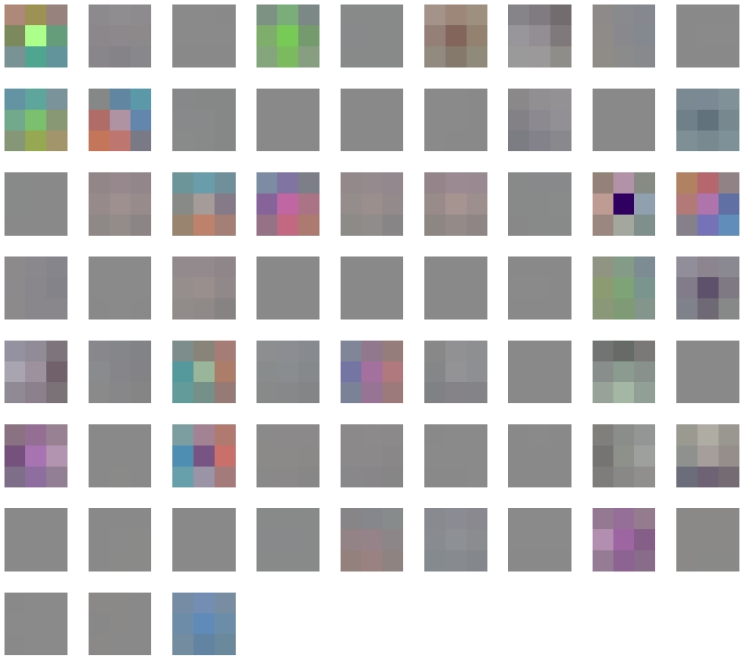}
    \caption{Filters learned in the first layer, when dropout is not used.}
  \end{subfigure}
  \caption{When dropout is not used, filters are not as vivid and more dead units are encountered, which is an indication of the presence of more noise.}
  \label{fig:new_conv1_weight}
\end{figure}

\begin{figure}[h!]
  \centering
  \begin{subfigure}[b]{0.48\linewidth}
    \includegraphics[width=\linewidth]{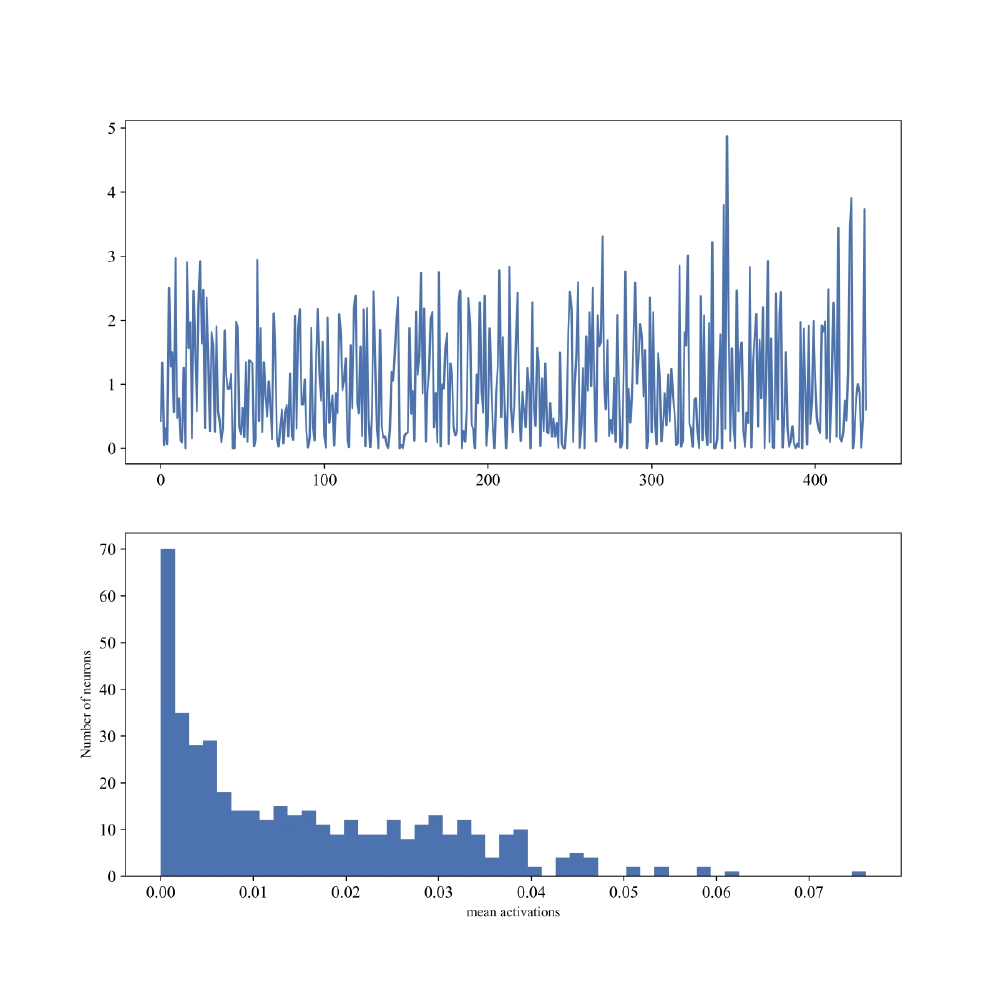}
    \caption{layer 13, using dropout.}
  \end{subfigure} ~
  \begin{subfigure}[b]{0.48\linewidth}
    \includegraphics[width=\linewidth]{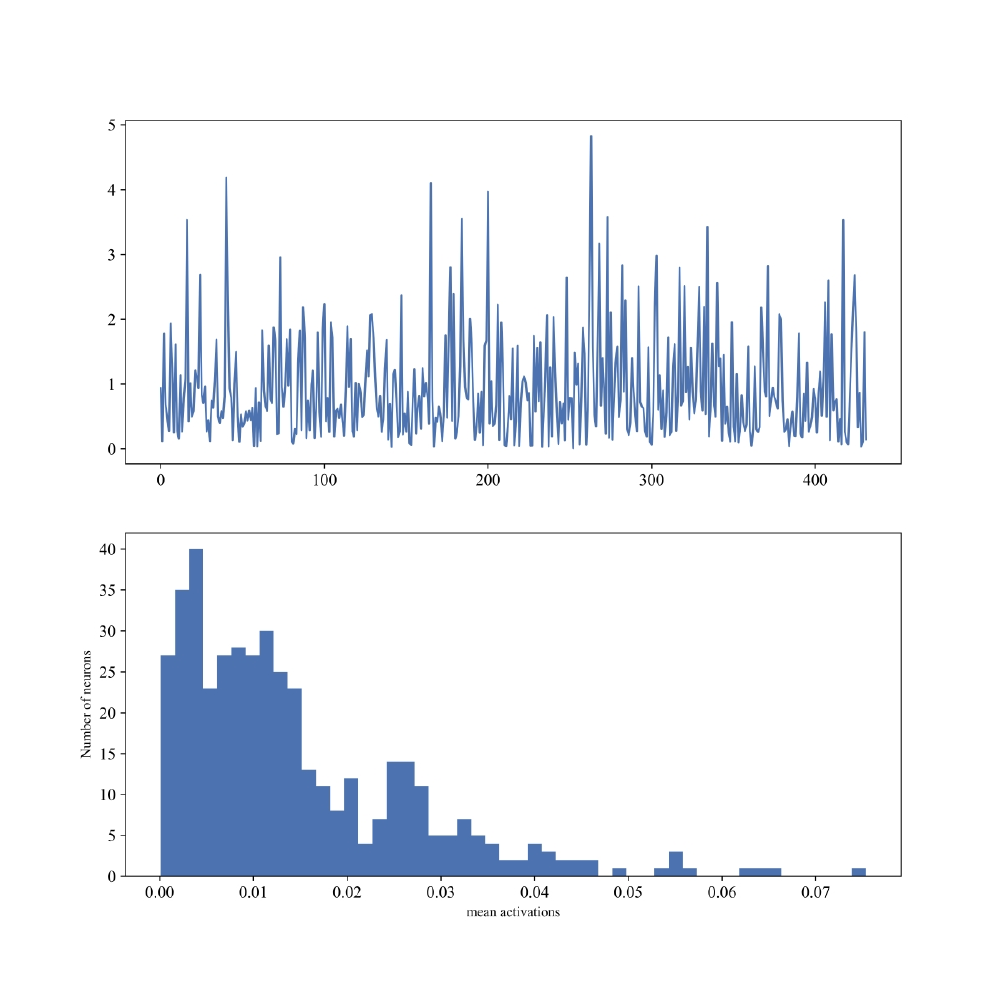}
    \caption{layer 13, using no dropout.}
  \end{subfigure}
  \caption{More neurons are activated around 0 when dropout is used which means more sparsity.}
  \label{fig:conv13_activations_and_hist}
\end{figure}

Using dropout has been an inseparable part of nearly all recent deep architectures, often considered as an effective regularizer. Dropout is also interpreted as an ensemble of several networks, which are trained on different subsets of the training data. It is believed that the regularization effect of dropout in convolutional layers is mainly influential for robustness to noisy inputs.
%Commonly, dropout has mostly been used on fully connected layers at the end of the network, to fight overfitting as a regularizer \cite{AlexKrizhevsky_imgnet_2012, Simonyan_VGG_2014, Szegedy_googlenet_2015}. However, it has been also used with Convolutional layers in some cases \cite{Springenberg_StrivingForSimplicity_2014,Huang_DenselyCNN_2016}, though less frequently. The reason is that it is considered to slow down the training process for lower layers, since the scale of back-propagated error decreases whenever it passes through a layer with dropout. Therefore, dropout has been used with fully connected layers more often \cite{Park_AnalysisOnDropout_2016}. %However we could not verify this behaviour for our case, probably because we have used BatchNormalization after each layer as well. 
To avoid overfitting, as instructed by \cite{Hinton_preventingCoAdapt_2012}, half of the feature detectors are usually turned off. However, this often causes the network to take a lot more to converge or even underfit, and therefore in order to compensate for that, additional parameters are added to the network, resulting in higher overhead in computation and memory usage. %Therefore either this technique was not used properly and thus prevented many architectures from taking advantage of its true potential. 
Towards better utilizing this technique, there have been examples such as \cite{Springenberg_StrivingForSimplicity_2014} that have used dropout after nearly all convolutional layers rather than only fully connected layers, seemingly to better fight overfitting of their large (at the time) network. Recently \cite{Huang_DenselyCNN_2016} also used dropout with all convolutional layers with less dropout ratio. Throughout our experiments and also in accordance to the findings of \cite{Park_AnalysisOnDropout_2016}, we found that applying dropout to all convolutional layers improves the accuracy and generalization power to some extent. This can be attributed to the behavior of neurons in different layers. One being the Dead ReLU issue, 
%One of the prevalent issues that have always been haunting networks based on rectified linear unit (ReLU) nonlinearity is the dead unit issue,
where a percentage of the whole network never gets activated and thus a substantial network capacity is wasted. The dead ReLU issue can be circumvented by using several methods (\eg, by using other variants of ReLU nonlinearity family \cite{He_PReLU_2015, Maas_RectifierNonlinearities_2013}). Another way to avoid this issue is to impose sparsity by randomly turning off some neurons, so that their contribution that may cause incoming input to the next neurons get negative, cancel out and thus avoid the symptom. The dropout procedure contributes to this by randomly turning off some neurons, and thus helps in avoiding the dead ReLU issue (see Figures \ref{fig:drpout_vs_nodrp_activations_firstlayer} and \ref{fig:new_conv1_fmap}). Additionally, for other types of nonlinearities incorporated in deep architectures, dropout causes the network to adapt to new feature combinations and thus improves its robustness (see Figure \ref{fig:new_conv1_weight}). It also improves the distributed representation effect \cite{Hinton_preventingCoAdapt_2012} by preventing complex co-adaptions. This can be observed in early layers, where neurons have similar mean values, which again is a desirable result, since early layers in a CNN architecture capture common features. This dropout procedure improves the generalization power by allowing better sparsity. This can be seen by visualizing the neurons activation in higher layers, which shows more neurons are activated around zero \cite{Park_AnalysisOnDropout_2016} (see Figure \ref{fig:conv13_activations_and_hist}).

%\begin{figure}
%  \includegraphics[width=0.6\linewidth]{drpall_new_conv1_plot1_1.jpg}
%  \caption{Showing Layer 1 neuron activations when dropout is used. }
%  \label{fig:drpout1}
%\end{figure}
%\begin{figure}
%  \includegraphics[width=0.6\linewidth]{nodrp_new_conv1_plot1_1.jpg}
%  \caption{Showing Layer 1 neuron activations when dropout is not used.}
%  \label{fig:drpout2}
%\end{figure}

%\begin{figure}[h!]
%  \centering
%  \begin{subfigure}[b]{0.49\linewidth}
%    \includegraphics[width=\linewidth]{drpall_new_conv1_plot1_1.pdf}
%    \caption{layer 1, using dropout.}
%  \end{subfigure}
%  \begin{subfigure}[b]{0.49\linewidth}
%    \includegraphics[width=\linewidth]{nodrp_new_conv1_plot1_1.pdf}
%    \caption{layer 1, using no dropout.}
%  \end{subfigure}
%  \caption{More neurons are frequently activated when dropout is used which means more filters are developed as apposed to when no dropout is used.}
%  \label{fig:conv1_activations_and_hist}
%\end{figure}

%\begin{figure}
%  \includegraphics[width=0.6\linewidth]{drpall_new_conv1.jpg}
%  \caption{Showing neuron activations at layer 1 when dropout is used.}
%  \label{fig:drpout_new_conv1_fmap}
%\end{figure}
%\begin{figure}
%  \includegraphics[width=0.6\linewidth]{nodrp_new_conv1.jpg}
%  \caption{Showing neuron activations at layer 1 when dropout is not used. When dropout is used, all neurons %are forced to take part and thus less dead neurons occur.}
%  \label{fig:nodrpout_enw_conv1_fmap}
%\end{figure}

%\begin{figure}
%  \includegraphics[width=0.6\linewidth]{drpall_new_conv1_weights.jpg}
%  \caption{Showing filters learned at layer 1 when dropout is used.}
%  \label{fig:drpoutall_new_conv1_weight}
%\end{figure}
%\begin{figure}
%  \includegraphics[width=0.6\linewidth]{nodrp_new_conv1_weights.jpg}
%  \caption{Showing filters learned at layer 1 when dropout is not used. When dropout is not used, filters are %not as vivid and more dead units are encountered which is an indication of the presence of more noise }
%  \label{fig:nodrpout_new_conv1_weight}
%\end{figure}

%\begin{figure}

%  \includegraphics[width=0.6\linewidth]{drpall_new_conv12_plot1_1.jpg}
%  \caption{layer 13, using dropout,More neurons are activated around 0 when dropout is used which means more %sparsity}
%  \label{fig:drpout_conv13_activations_and_hist}
%\end{figure}
%\begin{figure}
%  \includegraphics[width=0.6\linewidth]{nodrp_new_conv12_plot1_1.jpg}
%  \caption{layer 13, without any dropout}
%  \label{fig:nodrp_conv13_activations_and_hist}
%\end{figure}

\subsection{Simple Adaptive Feature Composition Pooling} \label{sec:poolX}
Max- and average-pooling are two methods for pooling operation that have been widely in use by nearly all architectures. While different architectures achieved different satisfactory results by using them and also many other variants \cite{Lee_CNN_Mixed_gated_2016}, we propose a variation, denoted as \textit{SAF-pooling}, which provides the best performance compared to conventional max-, average- pooling operations. \textit{SAF-pooling} is essentially the max-pooling operation carried out before a dropout operation. If we consider the input as an image, pooling operation provides a form of spatial transformation invariance, as well as reducing the computational complexity for the upper layers.
Furthermore, different images of the same class often do not share the same features due to the occlusion, viewpoint changes, illumination variation, and so on. Therefore, a feature which plays an important role in identification of a class in an image may not appear in different images of the same class. When max-pooling is used alongside (before) dropout, it simulates these cases by dropping high activations deterministically(since high activations are already pooled by max-pooling), acting as if these features were not present 
% Placing pooling after dropout, simulates the stochastic pooling operation, which turns off any neurons randomly and does not yield the same improvement as often. In our experiments we found that it performs inferior to its counterpart.
This procedure of dropping the maximum activations simulates the case where some important features are not present due to occlusion or other types of variations. Therefore, turning off high activations helps other less prominent but as important features to participate better. The network also gets a better chance of adapting to new and more diverse feature compositions. Furthermore in each feature-map, there may be different instances of a feature, and by doing so, those instances can be further investigated, unlike average pooling that averages all the responses feature-map-wise. This way, all those weak responses get the chance to be improved and thus take a more prominent role in shaping the more robust feature detector. 
This concept is visualized and explained in more details in supplementary material. %Therefore, the SAF-pooling operation needed two criteria, \ie, following a intuitive reasoning and a simple architecture (building block), which can be readily used. 
% Figure \ref{fig:New_pooling_layer_fig} demonstrates our idea.

\subsection{Final Regulation Stage}
While we try to formulate the best ways to achieve better accuracy in the form of rules or guidelines, they are not necessarily meant to be aggressively followed in all cases. These guidelines are meant to help achieve a good compromise between performance and the imposed overhead. Therefore, it is better to start by designing the architecture according to the above principles, and then altering the architecture gradually to find the best compromise between performance and overhead. %In order to better tune your architecture, try not to alter or deviate a lot from multiple guidelines at once. Work on one aspect at a time until the desired outcome is achieved. After all it’s all about the well balanced compromise between performance / imposed overhead according to one’s specific needs.

\begin{figure*}[ht!]
\begin{center}
\includegraphics[width=\linewidth]{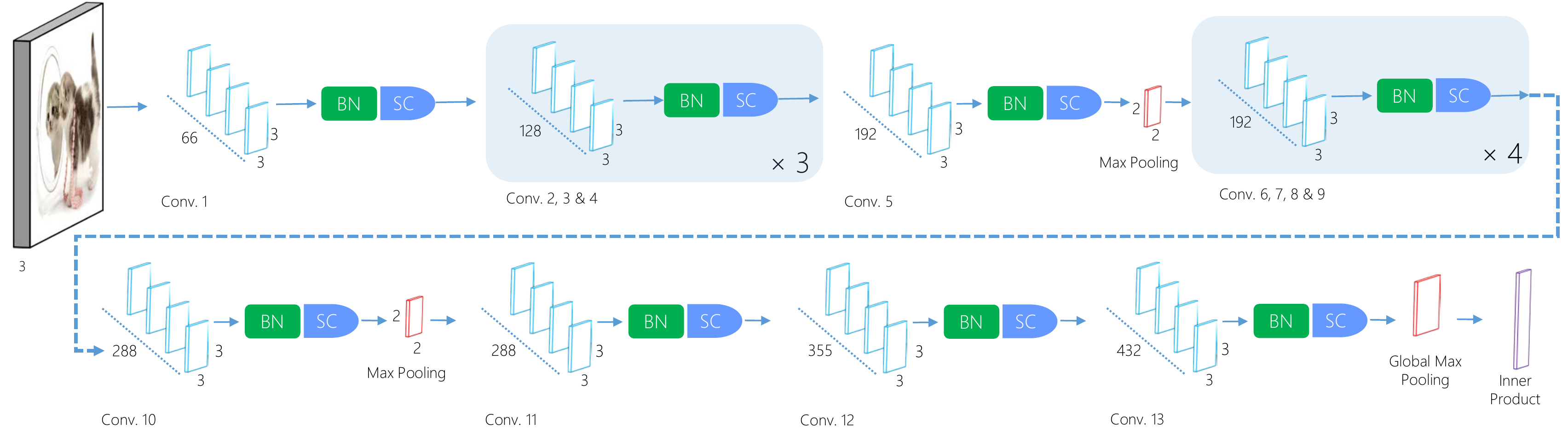}
\end{center}
 \caption{SimpNet base architecture with no dropout.}
\label{fig:The_arch_figure}
\end{figure*}

\section{SimpNet} \label{sec:the_arch}
Based on the principles mentioned above, we build a simple convolutional network with 13 layers. The network employs a homogeneous design utilizing $3\times 3$ filters for convolutional layers and $2\times 2$ filters for pooling operations. Figure \ref{fig:The_arch_figure} illustrates the proposed architecture. To design the network based on the above principles, we first specified the maximum depth allowed, to assess the architecture characteristics in comparisons with other deeper and wider counterparts. Following the first rule, we started by creating a thin but deeper architecture with 10 layers, then gradually increased its width to the point where no improvements were witnessed. The depth is then increased, and at the same time the last changes related to the width are reverted back. The experiments are reiterated and the width is increased gradually (related to the first rule and \textit{rapid prototyping}). During this process, we kept the width of all layers proportional and avoided any excessive allocation or lack thereof (\textit{balanced distribution scheme}). In order to manage this criteria, we designed layers in groups with similar characteristics, \ie, the same number of feature-maps and feature-map sizes (\textit{homogeneous groups}). To enable the extraction and preservation of as much information as possible, we used pooling operation sparingly, thus placing a pooling layer after each 5 layers. This lets the network to preserve the essential information with increased nonlinearity (\textit{maximum information utilization}). The exact locations where pooling should be placed can be easily determined through a few tests. Since groups of homogeneous layers are used, placing poolings after each group makes it an easy experiment. We used $3\times 3$ filters to preserve local information maximally, and also to utilize underlying performance optimizations in cuDNN library (\textit{maximum performance}). We then ran several experiments in order to assess different network characteristics, and in order to avoid sweeping unintentional details affecting the final result, we ran experiments in isolation, \ie, when an experiment is being run, all criteria are locked. Finally the network is revised in order to resolve specific internal limits, \ie, memory consumption and number of parameters (\textit{final regulation stage}). 
%Table 1 shows different architectures and their statistics, among which our architecture has the lowest number of parameters and operations. The extended list is provided in the appendix.

%\begin{table*}[h!]
%\caption{showing different architectures statistics}
%\begin{center}
%\begin{tabular}{lcccccc}
%\hline\hline
%Model & AlexNet \cite{?} &	GoogleNet \cite{?}	& ResNet152 \cite{?}  &	VGGNet16 \cite{?}  & 	NIN \cite{?}	&  Ours \\
%Param & 60M &	7M &	60M	& 138M	& 7.6M	& 5.4M\\
%OP &1140M  &	1600M  &	11300M  &	15740M &	1100M  &	652M\\
%Storage (MB) &217 & 51	 & 230 &	512.24  &	29	& 20\\
%\hline
%\end{tabular}
%\end{center}
%\label{tab:acc}
%\end{table*}

\section{Experimental Results} \label{sec:experiments_results}
In this section, several experiments are setup to show the significance of the introduced principles. To this end, results are generated without any hyperparameter tuning. These experiments are conducted using variants of our simple architecture. The overall architecture is the same as the one introduced in Section \ref{sec:the_arch} with slight changes in the architecture for two reasons, first to make it easier to conduct fair tests between different usecases, and second to provide more diversity in terms of the architecture essence. For example, we tried architectures with different depth and different number of parameters. Therefore, in each experiment, a different version of this network is used with an emphasis on the principle in question. Also, different cases within an experiment use the same architecture and optimization policy so that the test reflects only the changes required by the principle. We report the number of layers and parameters when evaluating each architecture, to see the effect of varying depth with fixed number of parameters for the same network. We present our experimental results in the following subsections, in which two different types of experiments are designed: first to evaluate the principles outlined in Section \ref{sec:intui}, and then to assess the performance of the proposed architecture based on those principles (\ie, SimpNet) compared to several leading architectures.

\subsection{Implementation Details}
The experiments concerning the design principles use CIFAR10 benchmark dataset, however, the second set of experiments are run on 4 major benchmarking datasets namely, CIFAR10, CIFAR100, SVHN and MNIST. CIFAR10/100 datasets \cite{Krizhevsky_LearningMultipleLayers_2009} include 60,000 color images, of which 50,000 belong to the training set and the rest are reserved for testing (validation), with 10 and 100 distinct classes, respectively. The classification performance is evaluated using top-1 error. The SVHN dataset \cite{Netzer_ReadingDigits_2011} is a real-world image dataset, obtained from house numbers in Google Street View images. It consists of 630,420 $32\times 32$ color images, 73,257 of which are used for training, 26,032 images are used for testing and the other 531,131 images are used for extra training. The MNIST dataset \cite{Lecun_GradientBased_CNN_1998} also consists of 70,000 $28\times 28$ grayscale images of handwritten digits, 0 to 9, with 60,000 images used for training and 10,000 for testing. 
We used Caffe framework \cite{Jia_Caffe_2014} for training our architecture and ran our experiments on a system with Intel Corei7 4790K CPU, 20 Gigabytes of RAM and NVIDIA GTX1080 GPU. 
\begin{table}[t!]
\centering
\caption{Gradual expansion of the number of layers.}
\label{tab:gradualexp}
\begin{tabular}{lcc}
%\hline
\textbf{Network Properties} & \textbf{Parameters} & \textbf{Accuracy (\%)}\\ \hline
%10 Layer & 302K   & 92.63 \\ %\hline
%11 Layer & 302K   & 92.56 \\ %\hline
%13 Layer & 302K   & 91.98 \\ \hline
Arch1, 8 Layers & 300K   & 90.21 \\ %\hline
Arch1, 9 Layers & 300K   & 90.55 \\ %\hline
Arch1, 10 Layers & 300K   & 90.61 \\ 
Arch1, 13 Layers & 300K   & 89.78 \\ \hline
\end{tabular}
\end{table}
\subsection{Examining the Design Intuitions}
The first set of experiments involve examining the design principles, ran on CIFAR10 \cite{Krizhevsky_LearningMultipleLayers_2009} dataset. 

\textbf{Gradual Expansion and Minimum Allocation}: Table \ref{tab:gradualexp} shows how gradually expanding the network helps obtaining better performance. Increasing the depth up to a certain point improves accuracy (up to 10 layers) and then after that it starts to degrade the performance. It is interesting to note that a 10-layer network is outperforming the same network with 13 layers. While in the next experiment(s), a deeper network is outperforming a shallower one with much fewer parameters. As already explained in the principles, namely in Subsections  \ref{sec:gradual_expansion_n_minumum_alloc} and \ref{sec:balanced_distribution}, one of the root causes that affects deeper architectures is an ill-distribution of processing capacity and more specifically the PLD/PLS issues, which can be seen in Table \ref{tab:gradualexp} and \ref{tab:minimum_alloc} respectively. 
%It should be noted that PLS and PLD issues are subjective criteria that should be taken into account for the same architecture.
These issues are indications of how a specific architecture manifests its needs, including more processing or representational capacity. 
%We tried to demonstrate at small scale what these phenomenas look like and why they happen, in order to have a glimpse of the underlying issue that prevents an efficient design or proper allocation.
It may be argued that the shallower architectures can still accommodate more capacity until they meet their true saturation point and, therefore, the difference in accuracy may be attributed to ill-distribution, rather than saturation (\ie, PLD/PLS). PLD/PLS are themselves manifestation of issues in proper distribution of processing units. Moreover, saturation is a process, which starts at some point and, as continues, it becomes more prominent. Our point is to give such impression and thus do not get into the specifics of finding the upper limit of an architecture saturation point. 
%As it is shown in several experiments results, even at smaller scale, this phenama can happen and be witnessed (Table \ref{tab:minimum_alloc}).
The networks used for \ref{tab:gradualexp} experiments are variants of SimpNet with different depths, denoted by \textbf{Arch1} in the table. More information concerning architectures and their overall topology is given in the supplementary material. 

In addition to the first test, Table \ref{tab:minimum_alloc} shows how a deeper network with fewer number of parameters works better than its shallower but wider counterparts. A deeper architecture can develop more interesting functions and thus the composition of many simpler functions can yield better inference, when properly capaciated. This is why despite the difference in the number of parameters, the deeper one is performing better. This experiment shows the importance of minimum allocation in the gradual expansion scheme. The networks used in \ref{tab:minimum_alloc} are variants of Arch1. Table \ref{tab:balanced_dist} demonstrates the results concerning balanced distribution of processing units throughout the architecture and how it excels against the ill-distributed counterpart. Networks used for \ref{tab:balanced_dist} use variants of \cite{hasanpour_letskeepit_2016}, which we call \textbf{Arch2} from which The first 10 layers are used for the 10-layer architecture and the 13-layer architecture is slimmed to have 128K parameters for the second test. 

\textbf{Maximum Information Utilization}: The Results summarized in the Table \ref{tab:maximum_utilization} show the effect of delayed pooling in an architecture, and demonstrate how utilizing more information in the form of bigger feature-maps can help the network to achieve higher accuracy. This table shows a variant of SimpNet with 13 layers and only 53K parameters, along with its performance with and without delayed pooling.
%That is when in the same architecture the pooling is applied further in the architecture as opposed to being applied earlier or more frequently.
We call this architecture \textbf{Arch3}. To assess the effect in question, one pooling operation is applied each time at different layers and everything else is kept fixed. L5, L3 and L7 refer to layer 5, 3 and 7, respectively. The results indicate an improvement compared to the initial design. 
%This means that only changing where the pooling should be applied (or its frequency) can easily affect how the architecture performs, thus demonstrating the effect of Maximum information utilization.

\begin{table}[t]
\centering
\caption{Shallow \vs~Deep (related to \textit{Minimum Allocations}), showing how a gradual increase can yield better performance with fewer number of parameters.}
\label{tab:minimum_alloc}
\begin{tabular}{lcc}
%\hline
\textbf{Network Properties} & \textbf{Parameters} & \textbf{Accuracy (\%)}\\ \hline
Arch1, 6 Layers &1.1M    & 92.18 \\ %\hline
Arch1, 10 Layers & 570K  & 92.23 \\ \hline
%Arch1, 13 Layers & 500K  & 92.42 \\ \hline
\end{tabular}
\end{table}

%\subsection{Balanced distribution Scheme}
\begin{table}[t]
\centering
\caption{Balanced distribution scheme is demonstrated by using two variants of SimpNet architecture with 10 and 13 layers, each showing how the difference in allocation results in varying performance and ultimately improvements for the one with balanced distribution of units.}
\label{tab:balanced_dist}
\begin{tabular}{lll}
%\hline
\textbf{Network Properties} & \textbf{Parameters} & \textbf{Accuracy (\%)}\\ \hline
Arch2, 10 Layers (wide end)         & 8M   & 95.19 \\ %\hline
Arch2, 10 Layers (balanced width)  & 8M   & 95.51 \\ %\hline
Arch2, 13 Layers (wide end)       & 128K & 87.20 \\ %\hline
Arch2, 13 Layers (balanced width) & 128K & 89.70 \\ \hline
\end{tabular}
\end{table}

\begin{table}[t]
\centering
\caption{The effect of using pooling at different layers. Applying pooling early in the network adversely affects the performance. }
\label{tab:maximum_utilization}
\begin{tabular}{lll}
%\hline
\textbf{Network Properties} & \textbf{Parameters} & \textbf{Accuracy (\%)}\\ \hline
Arch3, {L5} default          & 53K & 79.09 \\ %\hline
Arch3, {L3} early pooling    & 53K & 77.34 \\ %\hline
Arch3, {L7} delayed pooling & 53K & 79.44 \\ \hline
\end{tabular}
\end{table}

\begin{table}[t]
\centering
\caption{Effect of using strided convolution ($\dagger$) \vs~Max-pooling ($*$). Max-pooling outperforms the strided convolution regardless of specific architecture. First three rows are tested on CIFAR100 and two last on CIFAR10.} %implying the effect can be achieved in any architecture as long as pooling is used properly.
\label{tab:maximum_utilization_strided_vs_pooling}
\begin{tabular}{llll}
%\hline
\textbf{Network Properties}  & \textbf{Depth} &  \textbf{Parameters} & \textbf{Accuracy (\%)}\\ \hline
SimpNet$*$  & 13  &  360K & 69.28 \\ %\hline
SimpNet$*$  & 15   &   360K & 68.89 \\ %\hline
SimpNet$\dagger$ & 15  & 360K & 68.10 \\ %\hline
ResNet$*$  & 32  &  460K & 93.75 \\  
ResNet$\dagger$ & 32     & 460K & 93.46 \\ \hline
\end{tabular}
\end{table}

\textbf{Strided convolution vs. Maxpooling}: Table \ref{tab:maximum_utilization_strided_vs_pooling} demonstrates how a pooling operation regardless of architecture, dataset and number of parameters, outperforms the strided convolution. This proves our initial claim and thus nullifies the theory which implies 1. downsampling is what that matters, 2. strided convolution performs better than pooling. 
Apart from these, form the design point of view, strided convolution imposes more unwanted and unplanned overhead to the architecture and thus contradicts the principled design and notably gradual expansion and minimum allocation. To be more precise, it introduces adhoc allocation strategy, which in-turn introduces PLD issues into the architecture. This is further explained in supplementary material. 

\textbf{Correlation Preservation}: Table \ref{tab:correlationpresevation1} shows the effect of correlation preservation, and how a $3 \times 3$ architecture is better than its counterpart utilizing a bigger kernel size but with the same number of parameters. Table \ref{tab:correlationpresevation2} illustrates the same concept on different architectures and demonstrates how $1 \times 1$, $2\times 2$ and $3\times 3$ compare with each other at different layers. As it can be seen, applying $1 \times 1$ on earlier layers has a bad effect on the performance, while in the middle layers it is not as bad. It is also clear that $2\times 2$ performs better than $1\times 1$ in all cases, while $3\times 3$ filters outperform both of them. 

%\begin{table}[h!]
%\centering
%\caption{Demonstrating the effect of correlation preservation.}
%\label{tab:correlationpresevation1}
%\begin{tabular}{lll}
%%\hline
%\textbf{Network Properties} & \textbf{Parameters} & \textbf{Accuracy}\\ \hline
%$3\times 3$ & 5M & 87.9 (90.67) \\ %\hline
%$7\times 7$ & 5M & 87.14       \\ \hline
%\end{tabular}
%\end{table}
Here we ran several tests using different kernel sizes, with two different variants, one having 300K and the other 1.6M parameters. Using $7\times 7$ in an architecture results in more parameters, therefore in the very same architecture with $3\times 3$ kernels, $7 \times 7$ results in 1.6M parameters. Also, to decrease the number of parameters, layers should contain fewer neurons. Therefore, we created two variants of 300K parameter architecture for the $7\times 7$ based models to both demonstrate the effect of $7\times 7$ kernel and to retain the computation capacity throughout the network as close as possible between the two counter parts (\ie, $3\times3$ \vs~$7\times 7$). On the other hand, we increased the base architecture from 300K to 1.6M for testing $3\times 3$ kernels as well. Table \ref{tab:correlationpresevation1} shows the results for these architectures, in which the 300K.v1 architecture is an all $7\times 7$ kernel network with 300K parameters (\ie, the neurons per layer are decreased to match the 300K limit). The 300K.v2 is the same architecture with $7\times 7$ kernels only in the first two layers, while the rest use $3\times 3$ kernels. This is done, so that the effect of using $7\times 7$ filters are demonstrated, while most layers in the network almost has the same capacity as their counterparts in the corresponding $3\times 3$-based architecture. Finally, the $7\times 7$ 1.6M architecture is the same as the $3\times 3$ 300K architecture, in which the kernels are replaced by $7\times 7$ ones and thus resulted in increased number of parameters (\ie, 300K became 1.6M). Therefore, we also increased the $3\times 3$-based architecture to 1.6, and ran a test for an estimate on how each of these work in different scenarios. We also used a $5\times 5$ 1.6M parameter network with the same idea and reported the results. The network used for this test is the SimpNet architecture, with only 8 layers, denoted by \textbf{Arch4}. The networks used in Table \ref{tab:correlationpresevation2} are also variants of SimpNet with slight changes to the depth or number of parameters, denoted by Arch5.

\begin{table}[t]
\centering
\caption{Accuracy for different combinations of kernel sizes and number of network parameters, which demonstrates how correlation preservation can directly effect the overall accuracy.}
\label{tab:correlationpresevation1}
\begin{tabular}{lll}
%\hline
\textbf{Network Properties} & \textbf{Parameters} & \textbf{Accuracy (\%)}\\ \hline
Arch4, $3\times 3$ & 300K & 90.21 \\ %\hline
Arch4, $3\times 3$ & 1.6M & 92.14 \\ %\hline
Arch4, $5\times 5$ & 1.6M & 90.99 \\ %\hline
Arch4, $7\times 7$ & 300K.v1 & 86.09 \\
Arch4, $7\times 7$ & 300K.v2 & 88.57 \\
Arch4, $7\times 7$ & 1.6M & 89.22    \\ \hline
\end{tabular}
\end{table}

\begin{table}[t!]
\centering
\caption{Different kernel sizes applied on different parts of a network affect the overall performance, \ie, the kernel sizes that preserve the correlation the most yield the best accuracy. Also, the correlation is more important in early layers than it is for the later ones.}
\label{tab:correlationpresevation2}
\begin{tabular}{lll}
%\hline
\textbf{Network Properties} & \textbf{Params} & \textbf{Accuracy (\%)}\\ \hline
{\scriptsize Arch5,} {\tiny 13 Layers, $1\times 1$ \vs~$2\times 2$ (early layers)} & 128K & 87.71 \vs~88.50 \\ %\hline
{\scriptsize Arch5,} {\tiny 13 Layers, $1\times 1$ \vs~$2\times 2$ (middle layers)} & 128K & 88.16 \vs~88.51 \\ %\hline
{\scriptsize Arch5,} {\tiny 13 Layers, $1\times 1$ \vs~$3\times 3$ (smaller \vs~bigger end-avg)}    & 128K  & 89.45 \vs~89.60 \\ %\hline
{\scriptsize Arch5,} {\tiny 11 Layers, $2\times 2$ \vs~$3\times 3$ (bigger learned feature-maps)} & 128K  & 89.30 \vs~89.44 \\ \hline
\end{tabular}
\end{table}

%\subsubsection{Correlation preservation}
\textbf{Experiment Isolation}:
Table \ref{tab:isolation1} shows an example of invalid assumption, when trying different hyperparameters. In the following, we test to see whether $5\times 5$ filters achieve better accuracy against $3\times 3$ filters. Results show a higher accuracy for the network that uses $5\times 5$ filters. However, looking at the number of parameters in each architecture, it is clear that the higher accuracy is due to more parameters in the second architecture. The first architecture, while cosmetically the same as the second one, has 300K parameters, whereas the second one has 1.6M. A second example can be seen in Table \ref{tab:isolation2}, where again the number of parameters and thus network capacity is neglected. The example test was to determine whether placing $5\times 5$ filters at the beginning of the network is better than placing it at the end. Again here the results suggest that the second architecture should be preferred, but at a closer inspection, it is cleared that the first architecture has only 412K parameters, while the second one uses 640K parameters. Thus, the assumption is not valid because of the difference between the two networks capacities.

\begin{table}[t!]
\centering
\caption{The importance of experiment isolation using the same architecture once using $3\times 3$ and then using $5\times 5$ kernels.}
\label{tab:isolation1}
\begin{tabular}{lc}
%\hline
\textbf{Network Properties} & \textbf{Accuracy (\%)} \\ \hline
Use of $3\times3$ filters        & 90.21 \\ %\hline
Use of $5\times5$ instead of $3\times3$ & 90.99 \\ \hline
\end{tabular}
\end{table}
\begin{table}[t!]
\centering
\caption{Wrong interpretation of results when experiments are not compared in equal conditions (Experimental isolation).}
\label{tab:isolation2}
\begin{tabular}{lc}
%\hline
\textbf{Network Properties} & \textbf{Accuracy (\%)} \\ \hline
Use of $5\times 5$ filters at the beginning & 89.53 \\ %\hline
Use of $5\times 5$ filters at the end               & 90.15 \\ \hline
\end{tabular}
\end{table}

\textbf{SAF-pooling}: In order to assess the effectiveness of our intuition concerning Section \ref{sec:poolX}, we further ran a few tests using different architectures with and without \textit{SAF-pooling} operation on CIFAR10 dataset. We used SqueezeNetv1.1 and our slimmed version of the proposed architecture to showcase the improvement. Results can be found in Table \ref{tab:pooling_x_test}.
\begin{table}[t!]
\centering
\caption{Using \textit{SAF-pooling} operation improves architecture performance. Tests are run on CIFAR10.}
\label{tab:pooling_x_test}
\begin{tabular}{lll}
%\hline
\textbf{Network Properties} & \textbf{Accuracy (\%) With--without SAF Pooling}  \textbf{}\\ \hline
SqueezeNetv1.1 &  88.05(avg)--87.74(avg) \\ %\hline
%SqueezeNetv1.1 & Y & 88.06(avg) \\ %\hline
SimpNet &  94.76--94.68 \\ %\hline
%SipmNet & Y & 94.76 \\ 
\hline
\end{tabular}
\end{table}
\vspace{-3mm}

\subsection{SimpNet Results on Different Datasets}
SimpNet performance is reported on CIFAR-10/100 \cite{Krizhevsky_LearningMultipleLayers_2009}, SVHN \cite{Netzer_ReadingDigits_2011}, and MNIST \cite{Lecun_GradientBased_CNN_1998} datasets to evaluate and compare our architecture against the top ranked methods and deeper models that also experimented on these datasets. We only used simple data augmentation of zero padding, and mirroring on CIFAR10/100. Other experiments on MNIST \cite{Lecun_GradientBased_CNN_1998}, SVHN \cite{Netzer_ReadingDigits_2011} and ImageNet \cite{Russakovsky_ImageNet_2015} datasets are conducted without data-augmentation. In our experiments we used one configuration for all datasets and did not fine-tune anything except for CIFAR10. We did this to see how this configuration can perform with no or slightest changes in different scenarios.    

\subsubsection{CIFAR10/100}
%The CIFAR10/100 datasets \cite{Krizhevsky_LearningMultipleLayers_2009} include 60,000 color images of which 50,000 belong to training set and 10,000 are reserved for testing (validation). These images are divided into 10 and 100 classes respectively and classification performance is evaluated using top-1 error. 
Table \ref{tab:cifar} shows the results achieved by different architectures. We tried two different configurations for CIFAR10 experiment, one with no data-augmentation, \ie, no zero-padding and normalization and another one using data-augmentation, we achieved $95.26\%$ accuracy with no zeropadding and normalization and achieved $95.56\%$ with zero-padding. By just naively adding more parameters to our architecture without further fine-tuning or extensive changes to the architecture, we could easily surpass all WRN results ranging from 8.9M (with the same model complexity) to 11M, to 17M and also 36M parameters on CIFAR10/100 and get very close to its state-of-the-art architecture with 36M parameters on CIFAR100. This shows that the architecture, although with fewer number of parameters and layers, is still capable beyond what we tested with a limited budget of 5M parameters and, thus, by just increasing the number of parameters, it can match or even exceed the performance of much more complex architectures.  

\begin{table}[t!]
\centering
\caption{Top CIFAR10-100 results.}
\label{tab:cifar}
\begin{tabular}{lccc}
%\hline
\textbf{Method}                                                        & \textbf{\#Params}   & \textbf{CIFAR10}       & \textbf{CIFAR100}      \\ \hline
VGGNet(16L) \cite{Sergey_CIFAR10_OnTorch_2015}/Enhanced                                          & 138m       & 91.4 / 92.45  & -             \\ %\hline
ResNet-110L / 1202L \cite{He_ResNet_2015} *                                       & 1.7/10.2m & 93.57 / 92.07 & 74.84/72.18 \\ %\hline
SD-110L / 1202L \cite{Huang_DeepNN_StochDepth_2016} & 1.7/10.2m & 94.77 / 95.09 & 75.42 / -     \\ %\hline
WRN-(16/8)/(28/10) \cite{Zagoruyko_WRN_2016}                  & 11/36m    & 95.19 / 95.83 & 77.11/79.5  \\ %\hline
DenseNet \cite{Huang_DenselyCNN_2016}                                                                      & 27.2m      & 96.26         & 80.75         \\ %\hline
Highway Network \cite{Srivastava_HighwayNets_2015}                                                         & N/A          & 92.40         & 67.76         \\ %\hline
FitNet \cite{Romero_Fitnet_2014}                                                                           & 1M         & 91.61         & 64.96         \\ %\hline
FMP* (1 tests) \cite{Graham_FractionalMaxpooling_2014}                                  & 12M        & 95.50         & 73.61         \\ %\hline
Max-out(k=2) \cite{Goodfellow_MaxoutNetwork_2013}                                                          & 6M         & 90.62         & 65.46         \\ %\hline
Network in Network \cite{Lin_NIN_2013}                                                                     & 1M         & 91.19         & 64.32         \\ %\hline
DSN \cite{Lee_DeeplySupervisedNet_2015}                                              & 1M         & 92.03         & 65.43         \\ %\hline
Max-out NIN \cite{JiaRen_BatchNormMaxoutNIN_2015}                                           & -          & 93.25         & 71.14         \\ %\hline
LSUV \cite{Mishkin_AllYouNeedIsGoodInit_2016}                                & N/A          & 94.16         & N/A             \\ %\hline
SimpNet                                                                                            & 5.48M      & 95.49/95.56   & 78.08         \\ 
SimpNet                                                                                            & 8.9M      & 95.89   & 79.17         \\ \hline
\end{tabular}
\end{table}

\subsubsection{MNIST}
%The MNIST dataset \cite{Lecun_GradientBased_CNN_1998} consists of 70,000 28x28 grayscale images of handwritten digits 0 to 9, of which 60,000 are used for training and 10,000 are used for testing.
On this dataset, no data-augmentation is used, and yet we achieved the second highest score even without fine-tuning. \cite{Wan_Regularization_Using_DropConnect_2013} achieved state-of-the-art with extreme data-augmentation and an ensemble of models. We also slimmed our architecture to have only 300K parameters and achieved $99.73\%$. Table \ref{tab:MNIST} shows the result. %The full list is available in the supplementary material.% accuracy beating all previous larger and heavier architectures. Table 3 shows the current state-of-the-art results for MNIST.

\begin{table}[t!]
\centering
\caption{MNIST results without data-augmentation.}
\label{tab:MNIST}
\begin{tabular}{lc}
%\hline
\textbf{Method} & \textbf{Error rate} \\ \hline
Batch-normalized Max-out NIN \cite{JiaRen_BatchNormMaxoutNIN_2015}   & 0.24\%     \\ %\hline
Max-out network (k=2)  \cite{Goodfellow_MaxoutNetwork_2013}          & 0.45\%     \\ %\hline
Network In Network \cite{Lin_NIN_2013}                              & 0.45\%     \\ %\hline
Deeply Supervised Network  \cite{Lee_DeeplySupervisedNet_2015}       & 0.39\%     \\ %\hline
RCNN-96 \cite{Liang_RecurrentCNN_2015}                              & 0.31\%     \\ %\hline
SimpNet                                                   & 0.25\%     \\ \hline
\end{tabular}
\end{table}

\subsubsection{SVHN}
%The SVHN dataset \cite{Netzer_ReadingDigits_2011} is a %real-world image dataset, obtained from house numbers in %Google Street View images. It consists of 630,420 32x32 %color images of which 73,257 images are used for %training, 26,032 images are used for testing and the %other 531,131 images are used for extra training. 
Like \cite{Huang_DeepNN_StochDepth_2016, Goodfellow_MaxoutNetwork_2013, Lin_NIN_2013}, we only used the training and testing sets for our experiments and did not use any data-augmentation. Best results are presented in Table \ref{tab:SVHN}. Our slimmed version with only 300K parameters could achieve an error rate of $1.95\%$.
\begin{table}[h!]
\caption{Comparisons of performance on SVHN dataset.}\label{tab:SVHN}
\begin{center}
\begin{tabular}{lc}
%\hline
\textbf{Method} & \textbf{Error rate} \\ \hline
Network in Network\cite{Lin_NIN_2013}    &	2.35 \\
Deeply Supervised Net\cite{Lee_DeeplySupervisedNet_2015}	& 1.92 \\
ResNet\cite{He_ResNet_2015} (reported by \cite{Huang_DeepNN_StochDepth_2016} (2016)) &	2.01 \\
ResNet with Stochastic Depth\cite{Huang_DeepNN_StochDepth_2016}	&1.75\\
DenseNet\cite{Huang_DenselyCNN_2016} 	&1.79-1.59\\
Wide ResNet\cite{Zagoruyko_WRN_2016}  &	2.08-1.64\\
SimpNet	&1.648\\\hline

\end{tabular}
\end{center}
\end{table}

\subsubsection{Architectures with fewer number of parameters}
Some architectures cannot scale well, when their processing capacity is decreased. This shows the design is not robust enough to efficiently use its processing capacity. We tried a slimmed version of our architecture, which has only 300K parameters to see how it performs and whether it is still efficient. Table \ref{tab:Slimmed}  shows the results for our architecture with only 300K and 600K parameters, in comparison to other deeper and heavier architectures with 2 to 20 times more parameters. As it can be seen, our slimmed architecture outperforms ResNet and WRN with fewer and also the same number of parameters on CIFAR10/100.
%(DenseNet ro ezafe konim?)  
%
%\begin{table}[h!]
%	\begin{center}
%		\begin{tabular}{|l|c|c|c|c|c|c|c|c|}
%			\hline
%Model  &	Ours  & 	  &	  &	   &	 &	 &	  &    \\
%			\hline\hline
% Param  &	  &	&	&	 &	 &	 &		&  \\
%CIFAR10 &	  &	  &	  &	 &		&	& &	\\
%CIFAR10 &		&   &	 &	  &	  &	 & 		 & --- \\
%			\hline

\begin{table}[h!]
	\begin{center}
\caption{Slimmed version results on CIFAR10/100 datasets.}\label{tab:Slimmed}
		\begin{tabular}{lccc}
			%\hline
Model & Param & CIFAR10 &CIFAR100 \\ \hline
Ours & 300K	- 600K &93.25 - 94.03 & 68.47 - 71.74 \\
Maxout \cite{Goodfellow_MaxoutNetwork_2013} & 6M & 90.62 & 65.46 \\
DSN \cite{Lee_DeeplySupervisedNet_2015} & 1M &  92.03 & 65.43 \\
ALLCNN \cite{Springenberg_StrivingForSimplicity_2014} & 1.3M& 92.75& 66.29 \\
dasNet \cite{stollenga_dasnet_2014} & 6M & 90.78 & 66.22 \\
ResNet \cite{He_ResNet_2015} {\tiny(Depth32, tested by us)} & 475K &  93.22 & 67.37-68.95 \\
WRN \cite{Zagoruyko_WRN_2016}  & 600K &  93.15 & 69.11 \\
NIN \cite{Lin_NIN_2013}  & 1M	&91.19 &---\\ \hline
\end{tabular}
	\end{center}
	
\end{table}

\section{Conclusion} \label{sec:conclusion}
In this paper, we proposed a set of architectural design principles, a new pooling layer, detailed insight about different parts in design, and finally introduced a new lightweight architecture, SimpNet. SimpNet outperforms deeper and more complex architectures in spite of having considerably fewer number of parameters and operations, although the intention was not to set a new state-of-the-art, rather, showcasing the effectiveness of the introduced principles. We showed that a good design should be able to efficiently use its processing capacity and that our slimmed version of the architecture with much fewer number of parameters outperforms deeper and heavier architectures as well. Intentionally, limiting ourselves to a few layers and basic elements for designing an architecture allowed us to overlook the unnecessary details and concentrate on the critical aspects of the architecture, keeping the computation in check and achieving high efficiency, along with better insights about the design process and submodules that affect the performance the most. As an important direction for the future works, there is a calling need to study the vast design space of deep architectures in an effort to find better guidelines for designing more efficient networks.
%In this paper, we had to contend ourselves to a few configurations. We are still continuing our tests and would like to extend our work by experimenting on new applications and design choices especially using the latest achievements about deep architectures in the literature.%
\section*{Acknowledgment}\label{sec:acknowledgement}
\small{The authors would like to thank Dr. Ali Diba, CTO of Sensifai, for his invaluable help and cooperation, and Dr. Hamed Pirsiavash, Assistant Professor at University of Maryland-Baltimore County (UMBC), for his insightful comments on strengthening this work.}
%We would like to thank Dr. Ali Diba and  Dr. Hamed Pirsiavash for their helpful feedback, comments and cooperation.

\ifCLASSOPTIONcaptionsoff
  \newpage
\fi



%\bibliographystyle{IEEEtran}
\bibliographystyle{ieeetr}
{ \scriptsize
\bibliography{refs} 
}

\newpage


\section*{Supplementary material}

# Towards Principled Design of Deep Convolutional Networks: Introducing SimpNet (Supplementary Material)

Seyyed Hossein Hasanpour, Mohammad Rouhani, Mohsen Fayyaz, Mohammad Sabokrou and Ehsan Adeli

## S1. Extended Discussions

### A. Strided convolution vs. Max-pooling

Table VI (in the main paper) demonstrates how a pooling operation, regardless of architecture, dataset, and number of parameters, outperforms the strided convolution. This goes along our initial claim and also contradicts with the assumption which implies (1) Downsampling is what that matters; (2) Strided convolution performs better than pooling. Furthermore, from the design point of view, strided convolution imposes unwanted and unplanned overhead to the architecture and therefore contradicts the principled design and notably gradual expansion and minimum allocation. It introduces ad hoc allocation which in-turn causes PLD issues in the architecture. Suppose an optimal depth is $X$ with a given budget $Y$, using gradual expansion, one reaches to the depth $X$ using pooling operation, now replacing pooling with strided convolution, introduces $K$ layers to the architecture depth, ($K$ is the number of pooling layers). In the limited budget, the network will face PLD, since the balanced distribution is not optimal for the new depth. Furthermore, in order to cope with the PLD issue, the capacity needs to be increased by a large margin to accommodate $K$ new layers. The resulting architecture, then, ultimately under-performs against its counterpart using pooling operations, since strided convolution does not approximate a pooling layer and can only simulate the downsampling part of a pooling layer. This can be clearly seen from the obtained results. For the first part of the argument, it can be seen that the 13-layer architecture, which uses pooling with the same number of parameters , outperforms the strided version with 15 layers by a relatively large margin. The 15-layer version also outperforms the strided counterpart, although both show signs of PLD, the pooling based architecture performs much better. For the second part of the argument, a ResNet architecture with 32 layers is tested. ResNet uses strided convolutions and when it is altered to use pooling instead it performs much better. With the same depth, number of parameters and distribution scheme, the pooling-based architecture outperforms the strided convolution version. Therefore it is recommended not to use strided convolution instead of pooling.

### B. Simple Adaptive Feature Composition Pooling

Max- and average-pooling are the two methods for pooling operation that have been widely in use by nearly all architectures. While different architectures achieved different satisfactory results by using them and also many other variants

[1], we propose a variation, denoted as *SAF-Pooling*, which provides the best performance compared to conventional max-, average-pooling operations. *SAF-Pooling* is essentially the max-pooling operation carried out before a dropout operation.

### C. SAF-Pooling Intuition

If the input is considered as an image, pooling operation provides a form of spatial transformation invariance as well as reducing the computational complexity for the upper layers. However, by considering the input to the pooling layer as a series of different features, which is actually the case after appliance of several layers and feature representation possibly getting shrunk, it would be a mechanism which selects prominent features that describe an input identity. In this sense, an ideal pooling method is expected to preserve task-related information, while discarding irrelevant details. It is noteworthy that for the first intuition which considers input as an image, an overlapping pooling would be desired since the nonoverlapping pooling would lose considerably more spatial information in the input than its counterpart. However, when this assumption doesn't hold, nonoverlapped pooling would be a better choice. In our experiments, we noted that the nonoverlapping version outperforms the overlapped version by a relatively high margin, two intuitions may explain this improvement: (1) Due to the distributed representation and hierarchical nature of the network, and the fact that the neurons at higher level capture high-level and abstract features unlike the ones at earlier layers which are concerned with more low-level features and also (2) Deeper in the architecture, the essence of the input image changes and it is no more the same input image/representation rather, it is a feature-map, each pixel of which denotes a feature being present or not. Aafter some stages, the notion of image would not make sense at all, since the spatial dimension of the input may have shrunk so much that the identity inside image can no more be recognized (consider a $8 \times 8$ grid from a $32 \times 32$ pixel image of CIFAR10 dataset). Taking these two points into consideration, we are truly dealing with feature-maps in a network especially in mid to high layers. Thus, the second intuition seems more plausible, at least when pooling is carried out in the later layers as opposed to earlier ones, where the nonoverlapping may matter more because of more image like properties being still available. Furthermore, the first intuition may only apply to images in applications such as semantic segmentation and object detection based on current practices, and for other applications such as classification or other strategies in semantic

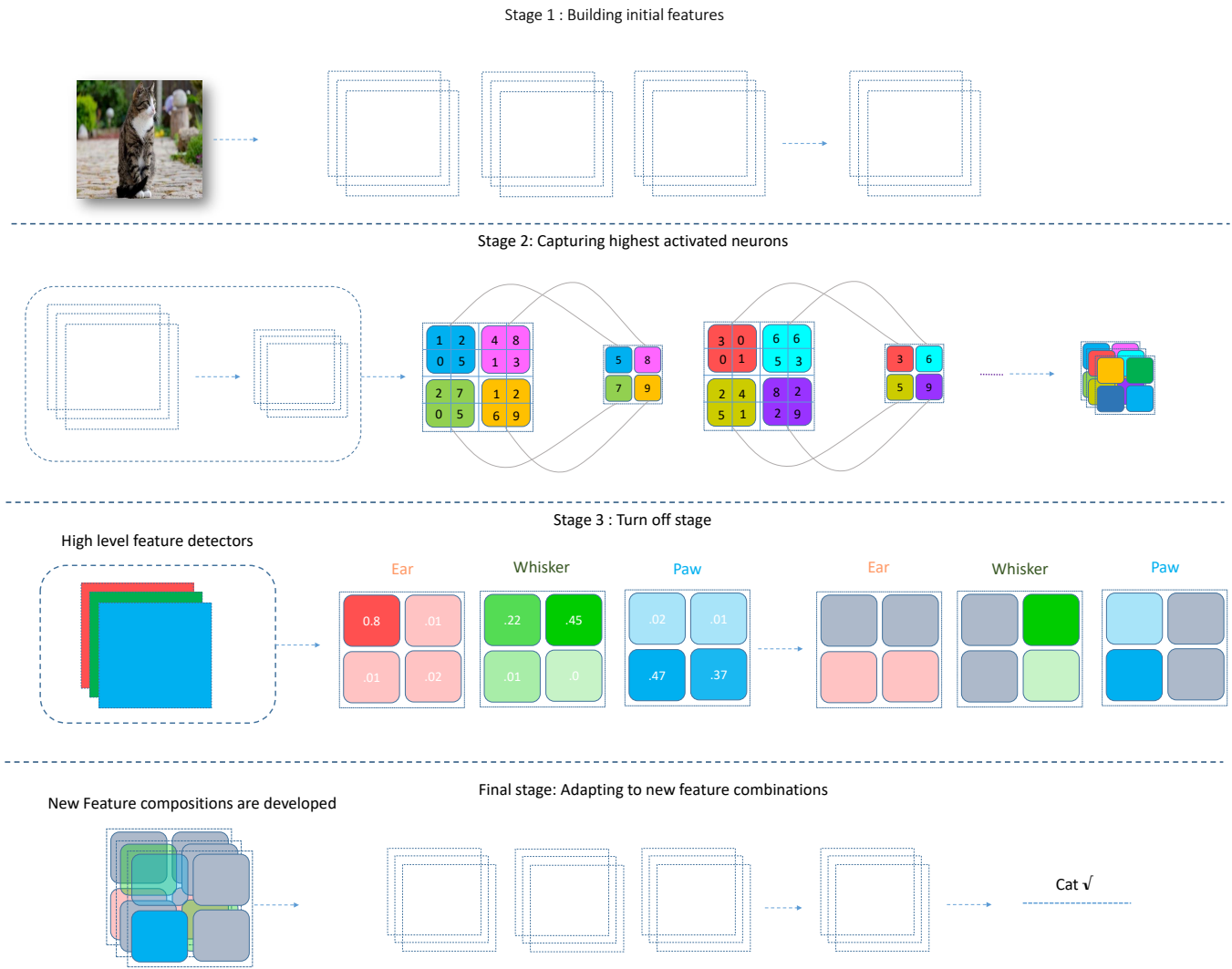

Fig. S1: SAF-Pooling: In this scheme, the network is forced to adapt itself to several different feature combinations through deterministically deactivating some feature responses, so that in cases some features are absent, due to occlusion, viewpoint changes, noise, *etc.*, the network can still carry out its task using other available features. The process happens in several stages. After some mid/high level features are available (Stage 1), most prominent features (responses) are pooled (Stage 2). Then, some of them are deterministically turned off. The network now has to adapt to the absence of some prominent features and therefore develop new feature compositions (Stage 3). After several iterations, less relevant (or irrelevant) features will be gradually replaced by more informative ones as weaker features are more frequently given the chance to be further developed. As the cycle continues, more feature compositions are developed which ultimately results in the network maximizing the capacity utilization, by creating more diverse feature detectors, as more neurons will be sparsely activated and independently learn patterns.

segmentation- or detection-based applications, it may not hold. Moreover, drastic loss of information that is claimed to be attributed to nonoverlapped pooling is usually caused by either drastic usage of pooling operations or frequent uses of them in the architecture, such as the cases that can be encountered in very deep architectures with several pooling layers, to take the computational burden as low as possible. Therefore, in the absence of such preconceptions/preconditions, the nonoverlapping pooling would not only pose an issue but would also be beneficial, as we experienced in our results as

well. Therefore, we followed the second intuition. Another form of pooling is average-pooling, but how does it fare with max-pooling in general? Average-pooling takes all activations in a pooling region into consideration with "equal contributions". This may downplay high activations as many low activations are averagely included. Max-pooling, however, only captures the strongest activations, and disregards all other units in the pooling region [2]. We noted that the usage of average-pooling specifically in early layers adversely affects the accuracy. However, in upper layers, it seems the difference between

the two methods (max and average) is much less severe. Furthermore, different images of the same class often do not share the same features due to the occlusion, viewpoint changes, illumination variation, and so on. Therefore, a feature which plays an important role in identification of a class in an image may not appear in different images of the same class. When max-pooling is used alongside (before) dropout, it simulates these cases by dropping high activations deterministically (since high activations are already pooled by max-pooling), acting as if these features were not present. Placing pooling after dropout, simulates the stochastic pooling operation, which turns off any neurons randomly and does not yield the same improvement as often. In our experiments, we found that it performs inferior to its counterpart. This procedure of dropping the maximum activations simulates the case where some important features are not present due to occlusion or other types of variations. Therefore, turning off high activations helps other less prominent but also as important features to also participate better and network gets a better chance of adapting to new and more diverse feature compositions. Furthermore, in each feature-map, there may be different instances of a feature, and by doing so, those instances can be further investigated. Unlike average pooling that averages all the responses feature-map wise, this way, all those weak responses get the chance to be improved and thus take a more prominent role in shaping the more robust feature detector. Moreover, in case those responses were false positives, they will be treated accordingly and suppressed in the back-propagation step, resulting in more robust feature detector development. Therefore, this results in more robust feature detectors and also better learning the class-specific characteristics [3], and feature composition adaptability by the network. Next, we present examples on how this improves the generalization power of the network greatly.

### D. SAF Pooling vs. Related Methods

SAF-Pooling operation looks very similar to the work of [3], however, there is a difference in the way the two operations are carried out. In [3], in the first step, the highest activations are set to zero. This scheme turns off the most informative features within the feature-map. However, in our case, we first pool the highest activations, and then randomly turn off some of them. Furthermore, we use the dropout ratio instead of a different probabilistic selection scheme as in [3]. This ensures that there are still enough features available not to hinder the speed of convergence. Also, at the same time by turning off some of the other highly activated values, the network is forced to learn more robust features to compensate for their absence, apart from that we avoid introducing new probabilistic scheme and making the procedure any further complex. Also [2] introduced a similar layer named *max-pooling dropout* which looks identical to the work presented in [3] with the difference that they use different probabilistic selection scheme. Wu *et al.* [2] also applied dropout before max-pooling and therefore performs some kind of stochastic pooling. On the contrary, the way *SAF-Pooling* is constructed and behaves is driven by the intuition we explained earlier and in the main paper (*i.e.*, *SAF-Pooling*), and also the fact that we are intentionally keeping everything simple. Figure S1 demonstrates our idea.

### E. Visualizing SAF-Pooling

In order to better demonstrate the underlying idea, and how it affects the performance of the network, we have run some experiments. In the following set of figures we show how the the notion of image is drastically changed through a network especially in higher layers and how this seemingly minor observation plays an important role in developing better intuitions about the network and how to improve its performance. This observation, in addition to the fact that poolings should not be applied early in the network, signifies that using overlapped pooling to keep image's spatial information does not make much sense, and if they are treated simply as feature-maps, it can lead to simple yet surprising ways of improvements (*i.e.*, SAF). In Figure S2, a head/neck detector is being demonstrated. Followed by a max-pooling operation, the most prominent responses are captured. This mechanism can then be viewed as a way to filter useful features for the rest of the pipeline and ultimately lead to the creation of robust feature compositions. Visualizations are created using DeepVis toolbox [4].

This simple observation has critical consequences in networks performance and the way it is constructed. Knowing what one deals with in an architecture, greatly helps in designing more intuitive and well-performing architectures. It seems logical when we are dealing with features, and we can simply filter prominent ones, we can enhance the feature composition part by explicitly creating new compositions on the fly. This hopefully will result in much better flexibility in the network, which would account for a lot of changes in the input (when some features are not present but others are). such intuitions mentioned so far, lead to the simple adaptive feature composition pooling. In Figure S3, one can see examples of how a network trained on CIFAR10, generalizes on hard to recognize samples. The network used Simple Adaptive feature composition pooling.

The following example also demonstrates how well the network has learned robust features to identify a class. We created a random image, that resembles several classes existing in CIFAR10, to see if the network can, identify several classes that make sense (*i.e.*, all 5 predictions need to be somehow relevant). Figure S4 shows the example image. Interestingly the network successfully identifies similar classes. Some features are valid for several classes and indeed network could use them to identify the corresponding identity. Not only a robust set of features need to have been learned, but also useful combinations of such features needed to exist. This helps the network to enjoy better generalization and avoid relying heavily on one feature as much as possible.

It should be noted that the image is not a dog, neither a cat nor a deer, or basically any specific creature/object. It was created in a way that it could express some features, thus resembling more than 1 class. The reason behind having such drawing, was initially to see if the network has a logical/reasonable predictions. For instance, it should predict similar classes rather than any random ones. If it does random predictions, it implies the network has failed to learned proper features to use. Please note that, by predictions, we refer to

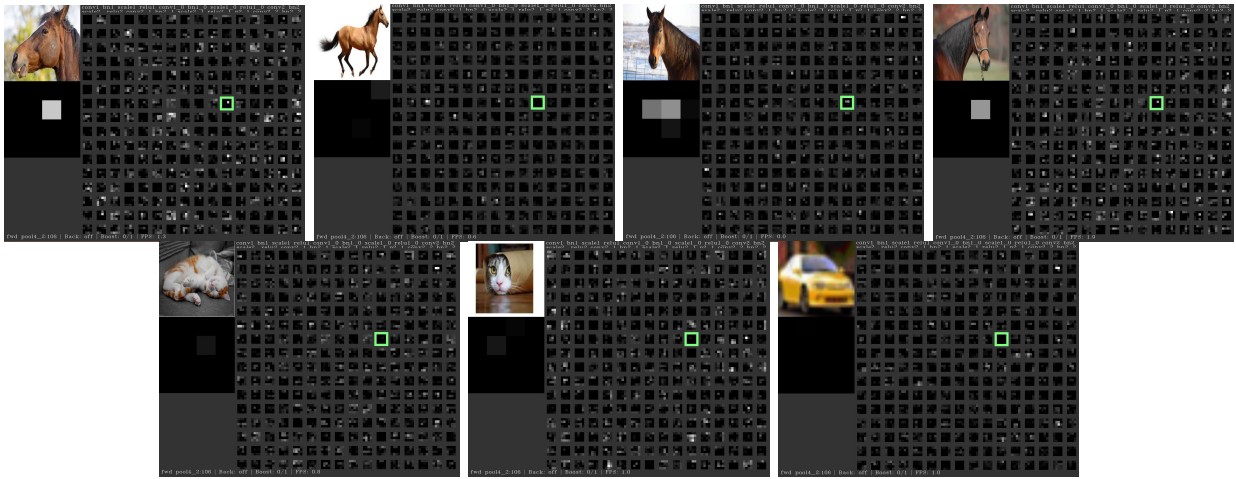

Fig. S2: Throughout the network, the notion of input changes into a feature-map, in which each response denotes the presence of a specific feature. Using Max-pooling, it is possible to filter out responses and capture the most prominent ones and use them to form interesting feature compositions later in the pipeline.

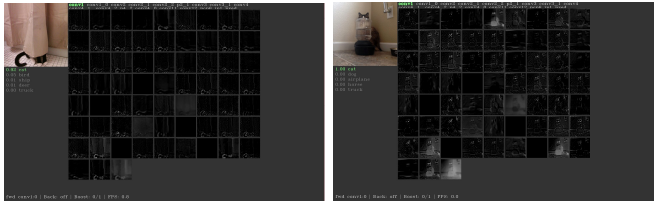

Fig. S3: At the absence of some features, network has learned to use other features and successfully and confidently recognize the cats.

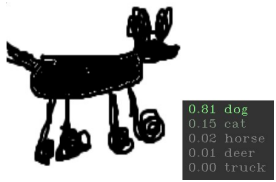

Fig. S4: Made up image to test network generalization and feature robustness.

the top 5 predictions. And by a logical/reasonable prediction, we mean, we would like the network to predict similar classes based on what exists in the image. The similarity between different classes are based on abstract attributes in those objects. A dog is different than a cat, so is a horse or a deer, but when we see a not clear image of a dog, we might mistake it for a cat, but we most probably never mistake it for a frog or a tin can! This kind of discrimination between different classes and yet be able to recognize and use inter-class features plays a critical role in our generalization capability. By looking at prominent features we can deduce the the actual object. As an example, by looking at a stickman drawing we can instantly say it is a depiction of a human, while a lot of coarse and fine-grained features related to humans and how they look are not even available, by looking at a very simplified drawing of a car we can identify it correctly. Now when we

are faced with such cases, that are not clear from the start, such as when we have not seen an object before hand ( the first encounter), or the shape/features are different what has already been observed, we use a similarity measure which is based completely on abstract attributes found in the objects we have thus far observed. Using this similarity measure we are able to say a picture resembles a cat and a bit less like a dog, but its definitely not a frog, a shark or an airplane. This information is crucial for improved generalization, because it needs to know prominent yet highly abstract features and then build on top of such information, the new similarity measure. Such information are vital for techniques such as Knowledge Distillation as well, which directly use them to train student networks. The more such information is accurate, the better the network performs on unseen data and the more useful it becomes for knowledge transfer. Such information is considered hidden knowledge which a network with holds and use to carry on its task unlike networks parameter values which are visible yet not interpretable normally. Therefore improving upon this subject can greatly benefit network performance, since we are implicitly providing the kind of information it mostly needs better. This is also a good indication of how well a network has learned and performs on unseen data. This is a very interesting observation, and clearly shows the network has learned exceptionally good features and their combinations to predict a class. This is again to point attention to the distributed nature of the neural network and thus treat feature-maps as they are and not simply as images. Having this insight lets utilizing the underlying learned features much efficiently.

The following images (Figures S5 and S6) also show how the network has developed robust features and how it is robust to drastic changes in the input. Note that this is trained on CIFAR10, without any fine-tuning, or heavy data-augmentation. It was simply trained on CIFAR10 with the same configuration explained before.

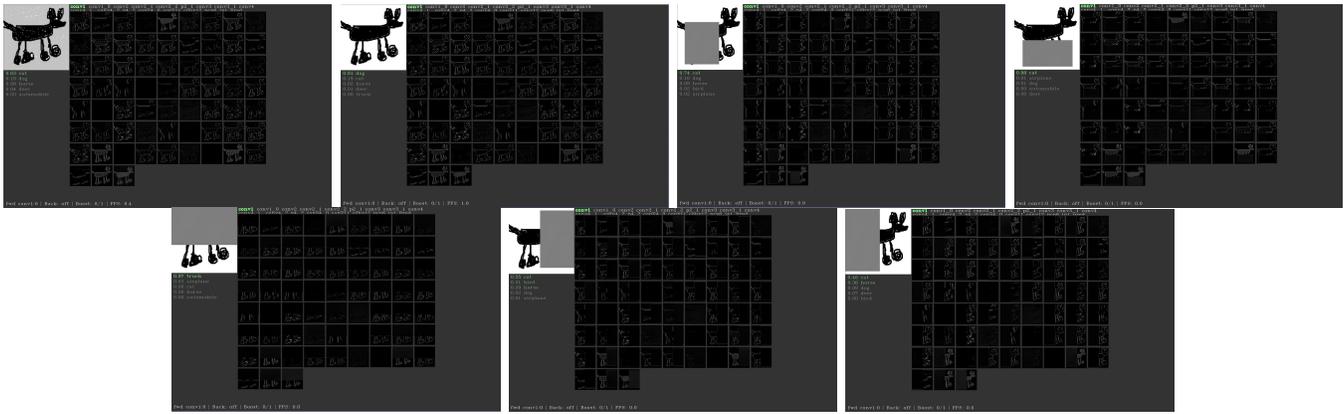

Fig. S5: The image that we created to see the network responses and see if its random or based on legitimately well learned features, here we can see the network could successfully list all classes that are close to each other can also can exists in the image

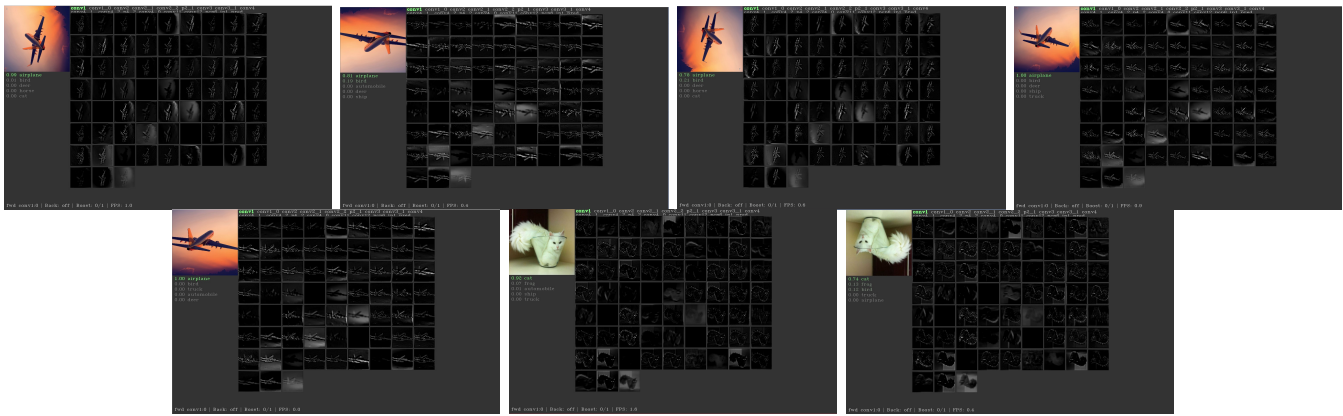

Fig. S6: Robust features are learned which is directly manifested in the very well generalization capability of the network

TABLE S1: Maximum performance utilization using Caffe, cuDNNv6, networks have 1.6M parameters and the same depth.

| Network Properties | 3×3 | 5×5 | 7×7 |
|---|---|---|---|
| Accuracy (higher is better) | 92.14 | 90.99 | 89.22 |
| Elapsed time(min)(lower is better) | 41.32 | 45.29 | 64.52 |

### F. Maximum Performance Utilization

Table S1 demonstrates the performance and elapsed time when different kernels are used. 3 × 3 has the best performance among the others.

### G. Correlation Preservation

SqueezeNet test on CIFAR10 *vs.* SimpNet (slim version), which use 3 × 3 convs everywhere.

## S2. EXPERIMENTS OF THE NETWORK ARCHITECTURES

For different tests we used a variety of network typologies based on main SimpNet (plain) architecture. Here we give general configuration concerning each architecture used in different tests reported in the main paper. The full implementation details along with other information is available at our Github repository. In what follows, C$x\{y\}$ stands for a convolution layer with $x$ feature-map, which is repeated $y$ times. When

TABLE S2: Correlation Preservation: SqueezeNet *vs.* SimpNet on CIFAR10. By *optimized* we mean, we added Batch-Normalization to all layers and used the same optimization policy we used to train SimpNet.

| Network | Params | Accuracy (%) |
|---|---|---|
| SqueezeNet1.1_default | 768K | 88.60 |
| SqueezeNet1.1_optimized | 768K | 92.20 |
| SimpNet_Slim | 300K | 93.25 |
| SimpNet_Slim | 600K | 94.03 |

TABLE S3: Gradual Expansion: Network Configurations.

| Network | Params | Config |
|---|---|---|
| 8 Layers | 300K | C41, C43, C70{4}, C85, C90 |
| 9 Layers | 300K | C44{2}, C57{4}, C75, C85, C96 |
| 10 Layers | 300K | C40{3}, C55{4},75, C85, C95 |
| 13 Layers | 300K | C30, C40{3}, C50{5}, C58{2}, C70, C90 |

$y = 1$, braces are not used. We use this form to demonstrate the overall architectures for the rest of the experiments.

### A. Gradual Expansion

The configurations used in table II are shown in table S3.

### B. Minimum Allocation

The configurations used in table III are shown in table S4.

2014.

TABLE S4: Minimum Allocation: Network Configurations

| Network | Params | Config |
|---------|--------|--------|
| 6 Layers | 1.1M | C64{2}, C128{2}, C256{2} |
| 10 Layers | 570K | C32, C48{2}, C96{7} |

TABLE S5: Balanced Distribution: Network Configurations. WD stands for Wide End and BW stands for Balanced Width.

| Network | Params | Config |
|---------|--------|--------|
| 10 Layers.WE | 8M | C64, C128{2}, C256{3}, C384{2},C512{2} |
| 10 Layers.BW | 8M | C150, C200{2}, C300{5},C512{2} |
| 13 Layers.WE | 128K | C19, C16{7}, C32{3}, C252,C300 |
| 13 Layers.BW | 128K | C19, C25{7}, C56{3},C110{2} |

TABLE S6: Maximum Information Utilization: Network Configuration.

| Network | Params | Config |
|---------|--------|--------|
| Arch | 53K | C6, C12{3}, C19{5}, C28{2}, C35, C43 |

TABLE S7: Correlation Preservation: Network Configurations

| Kernel Size | Params | Config |
|-------------|--------|--------|
| 3×3 | 300K | C41, C43, C70{4}, C85, C90 |
| 3×3 | 1.6M | C66, C96, C128{4}, C215, C384 |
| 5×5 | 1.6M | C53, C55, C90{4}, C110, C200 |
| 7×7.v1 | 300K | C20{2}, C30{4}, C35, C38 |
| 7×7.v2 | 300K | C35, C38, C65{4}, C74, C75 |
| 7×7 | 1.6M | C41, C43, C70{4}, C85, C90 |

TABLE S8: Correlation Preservation 2: Network Configurations.

| Kernel Size | Params | Config |
|-------------|--------|--------|
| 1×1.L1 →L3 | 128K | C21, C32{8}, C37, C48,c49, C50 |
| 2×2.L1 →L3 | 128K | C20, C32{8}, C38, C46{3} |
| 1×1.L4 →L8 | 128K | C20, C32{2}, C46{6}, C48{3}, C49 |
| 2×2.L4 →L8 | 128K | C20, C32, C33, C41{2}, C40{4}, C41, C43,44, C45 |
| 1×1.End | 128K | C19, C25{7}, C64, C53{2}, C108{2} |
| 3×3.End | 128K | C19, C25{7}, C26, C51{3},52 |
| 2×2.End | 128K | C19, C257, C91, C922 |
| 3×3.End | 128K | C20, C257, C643 |

TABLE S9: Using Different base architectures. All kernels are 3 × 3, if anything else is used it is specified inside parentheses (GP = Global Pooling).

| Arch1 | Arch2 | Arch3 | Arch4 | Arch5 |
|-------|-------|-------|-------|-------|
| Conv1 | Conv1 | Conv1 | Conv1 | Conv1 |
| Conv2 | Conv2 | Conv2 | Conv2 | Conv2 |
| Conv3 | Conv3 | Conv3 | Conv3 | Conv3 |
| Conv4 | Conv4 | Conv4 | Conv4 | Conv4 |
| Conv5 | Conv5 | Conv5 | Conv5 | Conv5 |
| Pool | Conv5 | Pool | Pool | Pool |
| Conv6 | Pool | Conv6 | Conv6 | Conv6 |
| Conv7 | Conv7 | Conv7 | Conv7 | Conv7 |
| Conv8 | Conv7 | Conv8 | Conv8 | Conv8 |
| Conv9 | Conv8 | Conv9 | Conv9 | Conv9 |
| Conv10 | Conv9 | Conv10 | Conv10 | Conv10 |
| Pool | Pool | Pool | Pool | Pool |
| Conv11 | Conv10 | Conv11 | Conv11 | Conv11 |
| Conv12 | Conv11(1x1) | Conv12 | Conv12 | Conv12 |
| Conv13 | Conv12(1x1) | Conv13 | Conv13 | Conv13 |
| GP | Pool | GP | GP | GP |
| | Conv13 | | | |
| | Pool | | | |

## C. Balanced Distribution

The configurations used in table IV are shown in table S5.

## D. Maximum Information Utilization

The configurations used in table V are shown in table S6.

## E. Correlation Preservation

The configurations used in table VII are shown in table S7.

Table S8 contains the configurations pertaining to the Table VIII .



\end{document}